\documentclass{article}
\PassOptionsToPackage{numbers, compress}{natbib}
\usepackage[preprint]{neurips_2026}

\usepackage[utf8]{inputenc} 
\usepackage[T1]{fontenc}    
\usepackage{hyperref}       
\usepackage{url}            
\usepackage{booktabs}       
\usepackage{amsfonts}       
\usepackage{nicefrac}       
\usepackage{microtype}      
\usepackage{xcolor}         

\usepackage{algorithm}
\usepackage{algorithmic}
\usepackage{wrapfig}
\usepackage{epsfig}
\usepackage{multirow}

\usepackage{threeparttable}
\usepackage{colortbl}
\usepackage{enumitem}
\usepackage{siunitx}
\usepackage{bm}
\usepackage{bbm}
\usepackage[export]{adjustbox}
\usepackage{listings}
\usepackage{pifont}
\usepackage{dblfloatfix}
\usepackage{lipsum}
\usepackage{subcaption}
\usepackage{pifont}
\usepackage{colortbl}
\usepackage{arydshln}
\hypersetup{colorlinks,linkcolor={red},citecolor={green},urlcolor={magenta}}
\usepackage{amsthm}
\usepackage{soul} 

\usepackage[accsupp]{axessibility}  

\usepackage{amsmath}
\usepackage{amssymb}
\usepackage{graphicx} 
\usepackage{caption}
\usepackage{enumitem}
\usepackage{wrapfig}  
\usepackage{multirow} 
\usepackage{listings} 
\usepackage{algorithm}

\theoremstyle{definition}

\newcommand{\eg}{\textit{e}.\textit{g}.}

\makeatletter
\newcommand{\thickhline}{%
    \noalign {\ifnum 0=`}\fi \hrule height 1pt
    \futurelet \reserved@a \@xhline
}
\definecolor{mygray}{gray}{.9}

\definecolor{LightGray}{rgb}{0.92,0.92,0.92}
\definecolor{yellow}{RGB}{244,238,0}
\definecolor{red}{RGB}{254,1,1}

\definecolor{mygray}{gray}{0.9}
\definecolor{codecomment}{RGB}{100,150,100}
\definecolor{codedefine}{RGB}{0,0,255}
\definecolor{codefunc}{RGB}{120,20,150}
\definecolor{codepro}{RGB}{50,100,200}

\usepackage{xspace}
\newcommand{\Ours}{SpecLoR\xspace}
\title{\Ours: Spectral Lookahead Rectification for Motion-Coherent Text-to-Video Generation}

\author{%
Xu Zhang$^{1}$, \quad Yu Lu$^{1}$, \quad Ruijie Quan$^{1}$, \quad Zhaozheng Chen$^{2}$, \quad Bohan Wang$^{2}$, \quad Yi Yang$^{1}\thanks{Corresponding Author: Yi Yang.}$ \\[0.5em]
  \small{$^1$ ReLER, College of Artificial Intelligence, Zhejiang University \quad $^2$ Huawei Central Research Institute} \\[0.5em]
  \small{\textbf{Project Page:} \url{https://xuzhang2.github.io/SpecLoR/}} 
}

\begin{document}

\maketitle

\begin{abstract}
Flow Matching has enabled robust text-to-video generation via latent ODE sampling. However, velocity approximation and numerical discretization errors inevitably accumulate, causing sampling trajectories to drift. Consequently, generated videos often suffer from severe spatiotemporal inconsistencies. Nevertheless, directly correcting these drifted, noisy latents is challenging: (i) timestep-dependent noise obscures reliable structural cues; (ii) spatial interventions risk disrupting intricate local geometry while incurring heavy computational costs. To address this, we propose Spectral Lookahead Rectification (\textbf{\Ours}), a plug-and-play inference method that bypasses noise via lookahead prediction, and circumvents spatiotemporal entanglement by shifting corrections to the frequency domain, where universal statistical priors of natural videos are readily available. First, during early sampling stages, \Ours looks ahead to estimate the clean latent $z_{t,0}$ and computes its 3D spatiotemporal spectrum. Next, \Ours rectifies the amplitude spectrum to match the prior, leaving the phase intact. Finally, the corrected state is re-noised to resume ODE integration. Experiments on Wan2.2 demonstrate that \Ours significantly reduces physical artifacts and enhances motion coherence across multiple benchmarks with minimal computational overhead (4 additional NFEs). 
\end{abstract}

\section{Introduction}
\label{sec:intro}

Text-to-video (T2V) generation$_{\!}$~\citep{kong2024hunyuanvideo,wan2025wan,liu2024sora,team2025kling,villegasphenaki,zhang2025waver,gao2025seedance,seed2026seedance2} has recently achieved remarkable milestones, largely driven by the transition toward the Flow Matching~(FM) paradigm$_{\!}$~\citep{lipmanflow,liuflow}. Despite great progress, current T2V models still suffer from spatiotemporal inconsistencies, \eg, duplicated limbs, floating objects, or physically implausible contacts, especially under complex motion and interaction dynamics.

\begin{figure*}[t!]
    \centering
    \includegraphics[width=\linewidth]{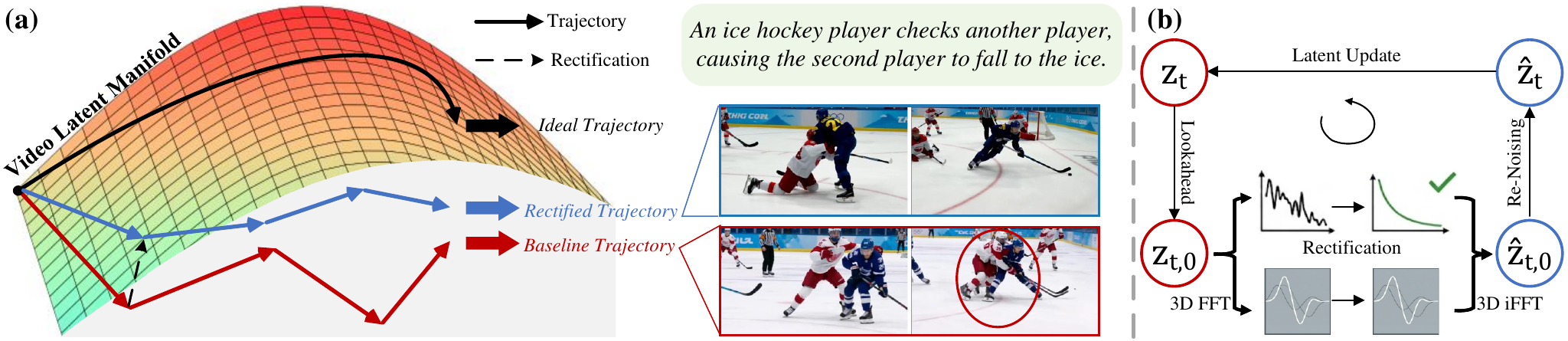}
    \caption{Concept and pipeline of \Ours. (a) Accumulated errors drift the Flow Matching trajectory from the ideal trajectory, yielding spatiotemporal inconsistencies. \Ours seamlessly anchors it back to a more valid trajectory. (b) At early timesteps, we project the noisy latent $z_t$ to a lookahead state $z_{t,0}$. We decouple its spatiotemporal spectrum, rectifying only the amplitude to a universal prior while preserving phase. The corrected state $\hat{z}_{t,0}$ is then re-noised to $\hat{z}_t$, resuming ODE integration.}
    \label{fig:fig1}
\end{figure*}

A primary cause of these inconsistencies lies in the \textit{sampling drift} during inference$_{\!}$~\citep{wu2024freeinit,xu2023restart,jang2026self}. In FM models, video generation proceeds by integrating a learned vector field using a numerical ODE solver. Because this estimated field cannot perfectly match the true continuous transport field, velocity approximation and numerical discretization errors inevitably accumulate across integration steps$_{\!}$~\citep{xu2023restart}. Consequently, the sampling trajectory gradually deviates from high-density regions of the data distribution, forcing intermediate latent states off the ideal generative path, as illustrated in Fig.\!~\ref{fig:fig1}a.
Existing inference-time methods attempt to mitigate this drift through trajectory exploration or perturbation. Search-based approaches$_{\!}$~\citep{singhalgeneral,oshimainference,he2025scaling} rely on rejection sampling or test-time selection to explore optimal sampling paths. However, these methods are often computationally prohibitive for high-dimensional video data. Noise initialization techniques$_{\!}$~\citep{wu2024freeinit,yuanfreqprior} tackle this issue at the trajectory’s starting point, but leave the overall ODE drift fundamentally unresolved. Meanwhile, perturbation-based methods$_{\!}$~\citep{xu2023restart,jang2026self} delve deeper into the intra-trajectory generation process by periodically injecting stochastic noise to dilute accumulated errors during inference. While mitigating accumulated errors, such blind perturbations lack a deterministic direction to steer the drifted trajectory back toward the natural video manifold, inadvertently introducing secondary distortions.

Rather than modifying the trajectory directly within the noisy latent space, where timestep-dependent noise severely obscures reliable spatiotemporal structural cues, we bypass this interference by projecting the latent onto a clean lookahead anchor. Crucially, we observe that trajectory drift manifests predominantly as \emph{spectral energy distortion}. Although the predicted lookahead state should ideally reconstruct the clean video latent, velocity approximation errors during early, high-noise steps distort its global frequency amplitude. Natural videos characteristically exhibit a power-law decay in their spatiotemporal spectrum, with the amplitude following a $1/f^{\alpha}$ distribution$_{\!}$~\citep{dong1995statistics}. However, trajectory deviations disrupt this natural prior, yielding abnormal spectral distributions that manifest as visible artifacts such as ghosting, duplicated structures, and unnatural motion. Since these distortions emerge precisely when the global motion structure is actively forming, correcting this drift at its inception proves highly effective. Consequently, early restoration of the natural spectral prior provides a robust structural constraint while preserving the instance-specific geometry encoded in the phase, successfully avoiding the complexity of altering the entire sampling trajectory.

Motivated by this insight, we introduce \textbf{Spectral Lookahead Rectification (\Ours)}, an inference-time sampling method that mitigates trajectory drift via spectral amplitude alignment. As illustrated in Fig.\!~\ref{fig:fig1}b, \Ours first projects the noisy latent $z_t$ onto its theoretical lookahead anchor, $z_{t,0}$, and extracts its spatiotemporal spectrum. Then, the amplitude spectrum is rectified to match this natural prior while preserving the intricate, geometry-encoding phase$_{\!}$~\citep{oppenheim1981importance}. Finally, the corrected latent $\hat{z}_{t,0}$ is re-noised back to the current timestep as $\hat{z}_t$ to resume ODE integration. By restoring the natural spectral magnitude during early sampling stages, \Ours acts as a reliable structural anchor. It steers the sampling trajectory back toward the natural video manifold, ensuring integration proceeds from a more physically valid state. Comprehensive evaluations on SOTA video generators demonstrate that \Ours effectively suppresses structural artifacts and enhances motion coherence across complex dynamics. Specifically, integrating \Ours into the advanced Wan2.2 framework substantially elevates visual fidelity and spatiotemporal consistency across multiple challenging benchmarks, while incurring minimal computational overhead (only 4 additional NFEs in a 40-step schedule).

Our contributions can be summarized as follows:

\begin{itemize}[leftmargin=*, nosep]
\item We identify spectral amplitude distortion as a key manifestation of sampling drift, providing a novel mechanistic insight into structural artifacts within Flow Matching-based video synthesis.

\item We introduce \textbf{\Ours}, a plug-and-play inference-time method that effectively calibrates the amplitude spectrum of the clean lookahead predictions using statistical video priors, seamlessly reintegrating this restored state to guide the drifted trajectory back onto the natural video manifold.

\item We reveal that spectral distortions emerge prominently during early sampling stages as global motion characteristics are established. By strategically confining our rectification to this critical window, \Ours acts as an efficient structural constraint, substantially enhancing spatiotemporal consistency in SOTA models (\eg, Wan2.2) with minimal overhead (only 4 additional NFEs).
\end{itemize}

\section{Related work}

\subsection{Large-Scale Text-to-Video Generation and Flow Matching}
Text-to-Video (T2V) generation has recently witnessed a paradigm shift from iterative Diffusion Models (DMs)$_{\!}$~\citep{ho2020denoising,rombach2022high,blattmann2023stable} to continuous-time Flow Matching (FM)$_{\!}$~\citep{lipmanflow,liuflow,albergo2023building,su2025theoretical} paradigms. By formulating generative modeling as an optimal transport problem, FM constructs straighter, more predictable probability paths than traditional diffusion, significantly accelerating convergence. Coupled with highly scalable Diffusion Transformers$_{\!}$~\citep{peebles2023scalable}, this framework has rapidly become the de facto backbone for recent state-of-the-art large video foundation models, including Sora$_{\!}$~\citep{liu2024sora}, Kling$_{\!}$~\citep{team2025kling}, HunyuanVideo$_{\!}$~\citep{kong2024hunyuanvideo,wu2025hunyuanvideo}, and Wan$_{\!}$~\citep{wan2025wan}. However, extending FM to high-fidelity, large-scale video synthesis heavily exacerbates a critical vulnerability during inference. The generative process relies on solving an Ordinary Differential Equation (ODE) via discrete numerical integration (\eg, Euler or UniPC$_{\!}$~\citep{zhao2023unipc}). Because predicting the true vector field for high-dimensional, spatiotemporally entangled video data is exceedingly difficult, especially in the high-noise early stages, the network inevitably produces significant velocity approximation errors during inference. As highlighted in recent studies on generative trajectory dynamics$_{\!}$~\citep{xu2023restart,jang2026self}, these step-by-step errors quickly accumulate into severe trajectory drift, irrevocably pulling the generated latents off the natural video manifold.

\subsection{Inference-Time Sampling and Trajectory Control}
To mitigate video generation artifacts, recent works have introduced various inference-time strategies$_{\!}$~\citep{chen2025temporal, fang2025inflvg, liu2025video, liu2025improving, luo2025enhance, nam2025optical,qiufreenoise, lu2024freelong, li2025longdiff, lee2025videoguide,chang2024how,hassan2025factorized,fei2025structure,gokmen2025ropecraft,zhu2025motionrag}. One line of research focuses on guidance modulation; for instance, CFG-Zero$_{\!}$~\citep{fan2025cfg} adjusts Classifier-Free Guidance$_{\!}$~\citep{ho2021classifier} to prevent color over-saturation. While effective for global contrast, it fails to resolve the progressive accumulation of sampling drift. Alternative approaches explore path search$_{\!}$~\citep{singhalgeneral,oshimainference, he2025scaling,na2024diffusion,rameshtest} via rejection sampling or test-time selection; however, these methods are often computationally expensive for high-dimensional video. Similarly, noise initialization techniques$_{\!}$~\citep{wu2024freeinit,yuanfreqprior} solely target the trajectory's initial state, leaving the subsequent ODE drift unmitigated. Another direction involves sampling interventions$_{\!}$~\citep{jang2026self, xu2023restart}. Restart Sampling$_{\!}$~\citep{xu2023restart}, for example, periodically injects noise to re-evaluate problematic segments, yet this stochasticity inadvertently introduces new errors into the trajectory. More recently, test-time optimization approaches such as FlowMo$_{\!}$~\citep{shaulov2025flowmo} enforce temporal coherence during inference. However, these gradient-dependent methods operate strictly in the spatial domain. Since global motion and fine geometric details are deeply entangled within the spatial latent, direct modifications often unintentionally blur semantic textures or incur prohibitive computational overhead. In contrast, SpecLoR shifts the structural constraint to the frequency domain, where the global amplitude spectrum and instance-specific phase details can be seamlessly decoupled and safely rectified without compromising visual fidelity.

\subsection{Spectral Dynamics and Natural Scene Statistics}
Analyzing visual signals in the frequency domain provides a mathematically rigorous perspective on generation quality. Natural Scene Statistics demonstrate that real-world videos exhibit a characteristic $1/f^\alpha$ power-law decay in their 3D spatiotemporal spectrum$_{\!}$~\citep{dong1995statistics}, which remains preserved within Video Autoencoder latent spaces$_{\!}$~\citep{wu2025improved,blattmann2023stable}. While recent image diffusion works$_{\!}$~\citep{lv2024fourier,si2024freeu,yang2025fam} explore frequency interventions, they primarily reweight 2D spatial features within noise-corrupted intermediate network states. Directly extending these to video fails because ODE trajectory drift fundamentally corrupts the macroscopic 3D energy distribution across frames. Similarly, previous video methods$_{\!}$~\citep{wu2024freeinit,yuanfreqprior} utilizing spatial filtering often operate on noisy states, erroneously conflating structural geometry with signal energy. \Ours is fundamentally distinct: rather than modulating noisy spatial features, it operates strictly in the 3D spatiotemporal frequency domain on a noise-free lookahead prediction. By explicitly enforcing the natural $1/f^\alpha$ amplitude prior, \Ours corrects trajectory drift at its root while leaving the delicate, geometry-encoding phase strictly intact.

\section{Spectral Perspective on Trajectory Drift}
\label{sec:3}
\begin{figure}[t!]
    \centering
    \includegraphics[width=\linewidth]{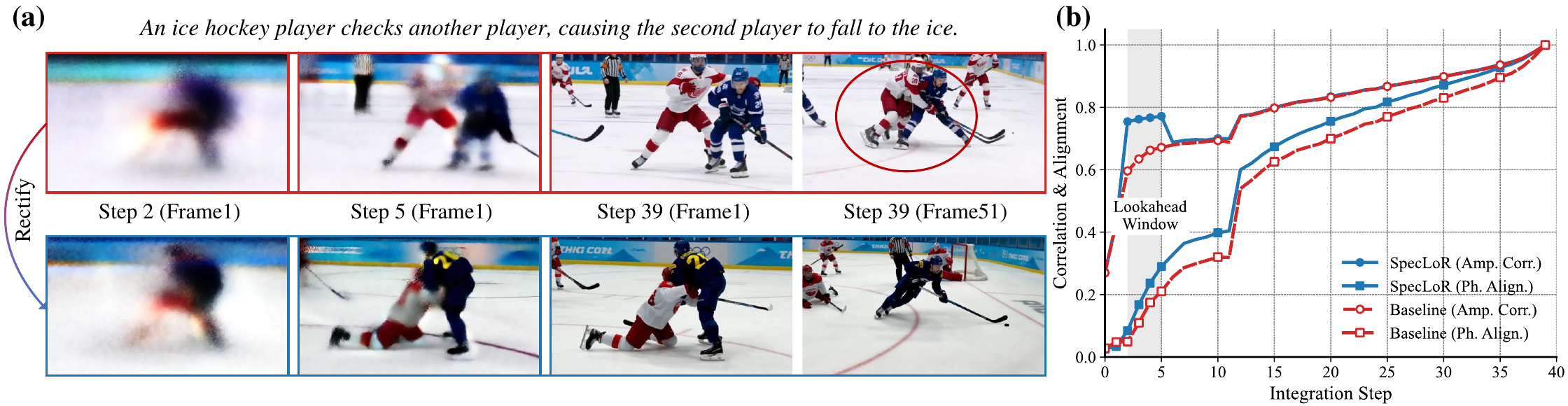}
    \caption{Trajectory drift and spectral rectification. (a) Visualizing intermediate lookahead predictions. Baseline (top): early structural ambiguity at Step 5 degrades into severe physical artifacts by Step 39 (red circles). \Ours (bottom) explicitly rectifies amplitude during early stages to establish a coherent global structure. (b) Quantitative tracking of frequency dynamics, measured by correlating intermediate steps against each method's final latent. Targeted amplitude rectification accelerates geometric phase convergence, fundamentally shifting the trajectory toward a valid endpoint.}
    \label{fig:fig2}
\end{figure}

In Flow Matching (FM), imperfect velocity predictions and numerical discretization inevitably introduce sampling trajectory drift. As errors accumulate, intermediate latent states deviate from the ideal generative path. Visualizing intermediate lookahead predictions (Fig.\!~\ref{fig:fig2}a) reveals a strict temporal dependency: severe structural artifacts at the end of generation (\eg, entangled instances at Step 39) fundamentally originate from structural ambiguities during early sampling stages.

Analyzing this phenomenon from a frequency perspective, we empirically observe that the macroscopic amplitude of the video latent, which governs global energy distribution, diverges significantly from natural priors during these early stages. Because amplitude provides the macroscopic energy foundation, this early drift permanently misguides the trajectory, forcing the structurally sensitive phase to converge into structurally incoherent states. This reveals a critical opportunity: mitigating trajectory drift does not necessitate full latent manipulation. In a pilot experiment, we rectified the amplitude spectrum exclusively during early ODE integration (Steps 2-5), leaving the phase untouched. To quantify the resulting dynamics (Fig.\!~\ref{fig:fig2}b), we calculate the correlation of intermediate latents against the final generated output at Step 40 of each respective method. We acknowledge that using each method's own final latent serves as a proxy target. Defining an absolute ``ground truth'' trajectory in text-to-video Flow Matching is theoretically ill-posed, as a single text condition maps to a highly multimodal distribution of valid videos rather than a singular deterministic endpoint. While our proxy lacks a strictly independent external reference, it effectively tracks the relative stabilization of the generative path. The curves reveal that this targeted amplitude intervention implicitly accelerates phase convergence towards this stabilized target, surpassing the baseline during the intervention window and maintaining this advantage. Crucially, this early constraint does not accelerate convergence to the flawed original destination; it fundamentally shifts the entire trajectory, enabling the model to arrive at a distinctly different and structurally superior endpoint.

\section{Method}

The central challenge in Flow Matching-based video generation is the accumulation of numerical and predictive errors, which inevitably forces the sampling trajectory off the natural video manifold. To address this, we introduce Spectral Lookahead Rectification (\Ours), a practical, highly efficient inference method. Classical signal processing theory$_{\!}$~\citep{oppenheim1981importance} establishes that the phase spectrum encodes delicate geometric details and is highly vulnerable to arbitrary interventions, which easily introduce severe structural artifacts (as empirically verified in \S\ref{subsec:diagnostic}). Concurrently, as quantitatively demonstrated by our spectral dynamics analysis (Fig.\!~\ref{fig:fig2}b), explicitly rectifying the macroscopic energy distribution (amplitude) effectively drives the geometric phase to converge. Grounded in this insight, our method mitigates trajectory drift via a structured four-stage pipeline, as illustrated in Fig.\!~\ref{fig:fig3}.

\subsection{Lookahead Projection}
\begin{figure}[t!]
    \centering
    \includegraphics[width=\linewidth]{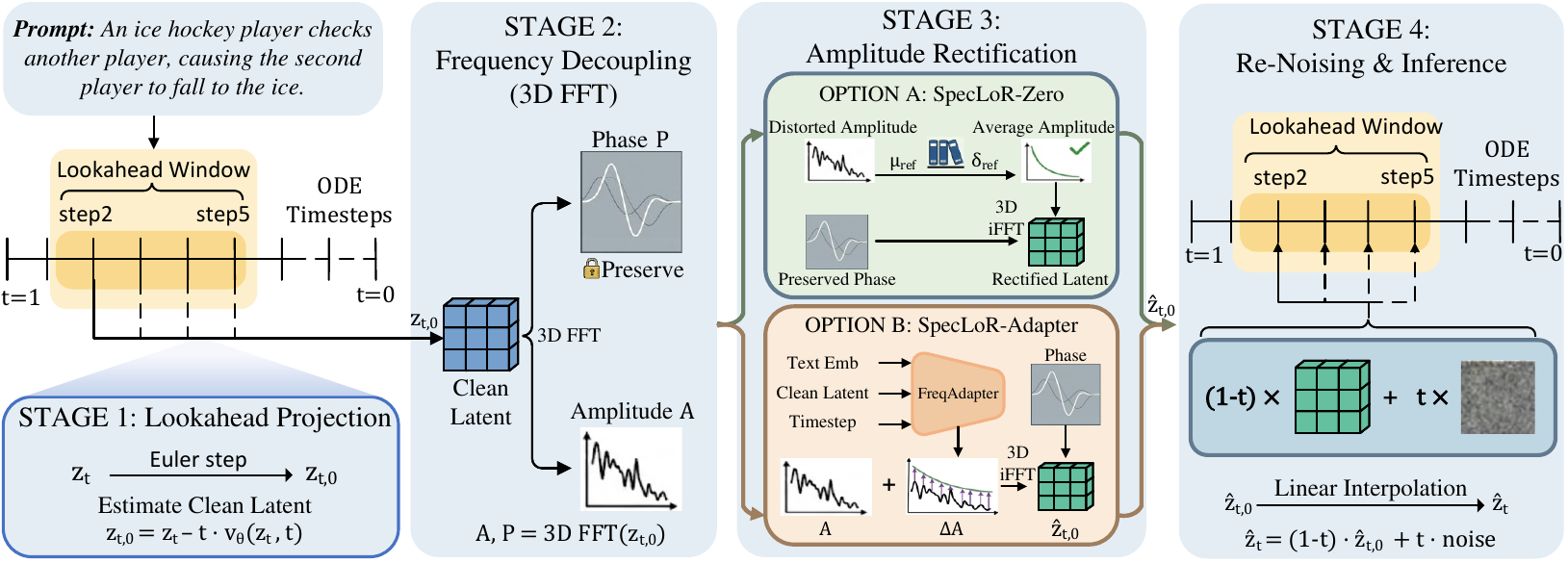}
    \caption{Pipeline of the proposed \Ours method. In Stage 1, the intermediate noisy latent $z_t$ is projected to a clean lookahead state $z_{t,0}$. Stage 2 explicitly decouples the spatiotemporal spectrum into amplitude and phase via 3D FFT. To correct trajectory drift, the corrupted amplitude is rectified using either a zero-cost global prior (Option A) or a context-aware adapter (Option B), while the detail-preserving phase is strictly locked. Finally, in Stage 4, the rectified spectrum is inverted and re-noised back to $\hat{z}_t$, providing a structurally sound anchor for the subsequent ODE integration.}
    \label{fig:fig3}
\end{figure}

Diagnosing trajectory drift directly within the intermediate noisy latent space $z_t$ is intractable due to the dominance of timestep-dependent noise. To bypass this, we utilize the current predicted vector field $v_\theta(z_t, t)$ to project the state into a noise-free space. As shown in Fig.\!~\ref{fig:fig3} (Stage 1), we estimate the clean latent $z_{t,0}$ via a single Euler step:
\begin{equation}
z_{t,0} = z_t - t \cdot v_\theta(z_t, t).
\end{equation}
This lookahead prediction $z_{t,0}$ acts as a diagnostic window, revealing early structural ambiguities before they solidify into physical artifacts. We explicitly constrain this operation to a specific ``Lookahead Window'' during the early high-noise regime (\eg, steps 2-5 in a 40-step schedule).

\subsection{Frequency Decoupling}
Directly modifying the spatial representation of $z_{t,0}$ is challenging because global motion structures and fine geometric details are tightly coupled. Any spatial intervention intended to correct this trajectory drift often unintentionally blurs semantic textures. To overcome this entanglement, we shift our intervention to the frequency domain (Fig.\!~\ref{fig:fig3}, Stage 2). Specifically, we apply a 3D Fast Fourier Transform (FFT) across the temporal and spatial dimensions ($T, H, W$) of $z_{t,0}$. This transformation explicitly decouples the macroscopic energy (Amplitude $\mathcal{A}$), which is susceptible to trajectory drift yet safe to rectify, from the highly vulnerable, geometry-encoding phase (Phase $\mathcal{P}$). By isolating and strictly preserving the phase $\mathcal{P}$, we establish a decoupled representation where the amplitude can be corrected without compromising instance-specific details or visual fidelity.

\subsection{Amplitude Rectification}

\noindent\textbf{\Ours-Zero (Global Prior).} According to Natural Scene Statistics$_{\!}$~\citep{dong1995statistics}, real-world videos consistently exhibit a $1/f^\alpha$ power-law spectral decay. However, trajectory drift prominently manifests as a severe deviation from this established prior. To rectify this, as depicted in (Fig.\!~\ref{fig:fig3} Option A), we introduce an asymmetric trust region derived from a universal natural prior ($\mu_{ref}, \sigma_{ref}$) pre-computed from roughly 2,000 randomly sampled real videos. We bound the current amplitude $\mathcal{A}_{curr}$ within this natural variance envelope to obtain the target amplitude $\mathcal{A}_{tgt}$:
\begin{equation}
\mathcal{A}_{tgt} = \operatorname{clip}(\mathcal{A}_{curr}, \; \gamma \cdot \mu_{ref}, \; \mu_{ref} + k \sigma_{ref}),
\end{equation}
where $\gamma$ (set to $0.6$) and $k$ (set to $3$) denote the lower and upper margin factors, respectively. We then apply a weighted update to obtain the final rectified amplitude:
\begin{equation}
    \mathcal{A}_{rect} = \mathcal{A}_{curr} + \lambda \cdot (\mathcal{A}_{tgt} - \mathcal{A}_{curr}),
\end{equation}
where $\lambda$ controls the intervention strength (empirically set to 0.5). This overhead-free structural anchor effectively stabilizes the trajectory, preventing physically implausible artifacts. 

\noindent\textbf{\Ours-Adapter (Context-Aware Refinement).} 
While the universal prior provides a robust foundation, ideal energy distributions vary across motion dynamics. To capture these instance-specific variations, we introduce the \Ours-Adapter (Fig.\!~\ref{fig:fig3}, Option B).

Constructed with Diffusion Transformer (DiT) blocks, the \Ours-Adapter leverages the predicted lookahead latent $z_{t,0}$, timestep embeddings $t_{emb}$, and text conditions to estimate an instance-specific residual amplitude map $\Delta\mathcal{A}$. During offline training on natural videos, we simulate the Flow Matching forward process on a ground-truth latent $z_0$ (with exact amplitude $\mathcal{A}_{gt}$) to sample an early-stage noisy latent $z_t$. After deriving the current lookahead amplitude $\mathcal{A}_{curr}$ from the predicted $z_{t,0}$, the adapter is optimized via a Mean Squared Error (MSE) loss in the logarithmic magnitude space:
\begin{equation}
    \mathcal{L} = \left\| \log(1 + \mathcal{A}_{curr} + \Delta\mathcal{A}) - \log(1 + \mathcal{A}_{gt}) \right\|_2^2.
\end{equation}
This log transformation compresses the spectrum's massive dynamic range caused by power-law decay. It prevents low-frequency dominance from overwhelming the loss, forcing the adapter to capture fragile high-frequency structural details while stabilizing training gradients.

Crucially, the base video model remains strictly frozen throughout training. At inference, the adapter yields the final context-aware rectified amplitude: $\mathcal{A}_{rect} = \mathcal{A}_{curr} + \Delta\mathcal{A}$.

\subsection{Re-Noising \& Inference}

Once the amplitude is rectified, we combine $\mathcal{A}_{rect}$ with the securely preserved phase $\mathcal{P}$. A 3D Inverse FFT (iFFT) projects them back into the latent space, yielding a purified, structurally rectified clean anchor $\hat{z}_{t,0}$. To integrate this corrected state back into the ODE solver (Fig.\!~\ref{fig:fig3}, Stage 4), we re-noise $\hat{z}_{t,0}$ back to the exact current timestep $t$ via linear interpolation with a noise term $\epsilon$:
\begin{equation}
\hat{z}_t = (1 - t) \cdot \hat{z}_{t,0} + t \cdot \epsilon.
\end{equation}
\noindent\textbf{Deterministic Continuation:} By setting $\epsilon = \epsilon_{init}$, we recycle the exact initial noise map to serve as a strict control. By fixing the randomness, we ensure that any shift in the trajectory is exclusively driven by our \Ours intervention. This perfectly preserves the spatiotemporal dynamics of the baseline, providing a strictly controlled visual environment that directly proves our method's effectiveness in eliminating physically implausible motion and structural artifacts.

\noindent\textbf{Stochastic Contraction:} By setting $\epsilon \sim \mathcal{N}(0, I)$, we inject freshly sampled stochastic noise. Drawing on the error-contraction properties of generative processes$_{\!}$~\citep{xu2023restart}, introducing stochasticity helps contract accumulated ODE discretization errors. Since \Ours already anchors the rectified macroscopic physical trajectory, this fresh noise optimizes local details without causing spatiotemporal collapse, yielding a globally more physically coherent video distribution.

\section{Experiments}

To comprehensively evaluate the effectiveness of the proposed \Ours, we conduct extensive experiments focusing on motion coherence, overarching visual quality, and computational efficiency.

\subsection{Experiment Setup}

\noindent\textbf{Benchmarks.} We meticulously evaluate our method across two rigorous benchmarks tailored for physical and temporal stability: (1) \textbf{Dynamic-Bench}$_{\!}$~\citep{jang2026self} to assess structural stability under challenging, multi-object interactions and complex human motions; and (2) \textbf{VideoJAM-Bench}$_{\!}$~\citep{chefer2025videojam} to target the joint appearance-motion coherence of generated dynamics.

\noindent\textbf{Evaluation Metrics.} To ensure a multi-dimensional assessment, we categorize our metrics as follows:
\begin{itemize}[leftmargin=*, nosep]
    \item \textit{Visual \& Physical Integrity:} Evaluated via the Visual Quality, Physical Consistency, and T2V Alignment scores from VideoScore2$_{\!}$~\citep{he2025videoscore2}, alongside the overall Consistency from VBench$_{\!}$~\citep{huang2024vbench}.
    \item \textit{Temporal Dynamics:} Assessed using Motion Smoothness and Dynamic Degree from VBench.
    \item \textit{Computational Efficiency:} Strictly tracked via the Number of Function Evaluations (NFE).
\end{itemize}

\noindent\textbf{Implementation Details.} Built upon \texttt{Wan2.2-A14B T2V}$_{\!}$~\citep{wan2025wan} (default 40 NFEs), \Ours operates exclusively during the early ``Lookahead Window'' (Steps 2-5). This targeted, inference-only intervention inherently bounds computational overhead, requiring only 44 NFEs. Experiments are conducted on 910B NPU with 60GB of memory, taking approximately 20 minutes to generate a single video.
\begin{table}[t]
  \centering
  \caption{Quantitative comparison. Best scores are in \textbf{bold} and second best are \underline{underlined}.}
  \label{tab:main_results}
  \resizebox{\textwidth}{!}{
\begin{tabular}{lccc cc cc}
    \toprule
    & \multicolumn{3}{c}{VideoScore2} & \multicolumn{2}{c}{VBench} & \multicolumn{2}{c}{Efficiency} \\
    \cmidrule(r){2-4} \cmidrule(lr){5-6} \cmidrule(l){7-8}
    Method & T2V Align.$\uparrow$ & Phys. Cons.$\uparrow$ & Visual Qual.$\uparrow$ & Consistency$\uparrow$ & Motion Smooth.$\uparrow$ & NFE$\downarrow$ & Time$\downarrow$ \\
    \midrule
    \multicolumn{8}{c}{\textit{Results on VideoJAM-Bench$_{\!}$~\citep{chefer2025videojam}}} \\
    \midrule
    Wan2.2 T2V (UniPC) & 4.062 & 3.237 & 3.873 & 0.935 & 0.972 & 40  &  \\
    + NFE$\times$1.5 & 4.046 & 3.260 & 3.834 & 0.935 & 0.974 & 60 & 1.5$\times$  \\
    + FlowMo$_{\!}$~\citep{shaulov2025flowmo} & 4.044 & 3.179 & 3.805 & 0.939 & 0.974 & 40 & 1.7$\times$  \\
    + SRVS$_{\!}$~\citep{jang2026self} & 4.067 & 3.341 & 3.926 & \textbf{0.943} & \textbf{0.979} & 60 & 1.5$\times$  \\
    + \Ours-Zero (Ours) & \underline{4.118}$_{\textcolor{gray}{\pm0.013}}$ & \textbf{3.448}$_{\textcolor{gray}{\pm0.018}}$ & \textbf{3.939}$_{\textcolor{gray}{\pm0.017}}$ & 0.939$_{\textcolor{gray}{\pm0.002}}$ & \underline{0.977}$_{\textcolor{gray}{\pm0.004}}$ & 44 & 1.1$\times$ \\
    + \Ours-Adapter (Ours) & \textbf{4.126}$_{\textcolor{gray}{\pm0.011}}$ & \underline{3.441}$_{\textcolor{gray}{\pm0.019}}$ & \underline{3.927}$_{\textcolor{gray}{\pm0.014}}$ & \underline{0.940}$_{\textcolor{gray}{\pm0.003}}$ & \underline{0.977}$_{\textcolor{gray}{\pm0.002}}$ & 44 & 1.1$\times$  \\
    \midrule
    \multicolumn{8}{c}{\textit{Results on Dynamic-Bench$_{\!}$~\citep{jang2026self}}} \\
    \midrule
    Wan2.2 T2V (UniPC) & \underline{3.605} & 2.687 & 3.745 & 0.909 & 0.974 & 40  &   \\
    + NFE$\times$1.5 & 3.541 & 2.898 & 3.688 & 0.912 & 0.974 & 60 & 1.5$\times$  \\
    + FlowMo$_{\!}$~\citep{shaulov2025flowmo} & 3.585 & \textbf{3.086} & 3.745 & 0.911 & 0.975 & 40 & 1.7$\times$  \\
    + SRVS$_{\!}$~\citep{jang2026self} & 3.584 & 2.993 & 3.800 & \textbf{0.917} & \textbf{0.981} & 60 & 1.5$\times$  \\
    + \Ours-Zero (Ours) & \textbf{3.651}$_{\textcolor{gray}{\pm0.012}}$ & 2.993$_{\textcolor{gray}{\pm0.012}}$ & \underline{3.810}$_{\textcolor{gray}{\pm0.014}}$ & \underline{0.915}$_{\textcolor{gray}{\pm0.005}}$ & \underline{0.980}$_{\textcolor{gray}{\pm0.003}}$ & 44 & 1.1$\times$  \\
    + \Ours-Adapter (Ours) & 3.588$_{\textcolor{gray}{\pm0.014}}$ & \underline{3.058}$_{\textcolor{gray}{\pm0.021}}$ & \textbf{3.864}$_{\textcolor{gray}{\pm0.013}}$ & 0.914$_{\textcolor{gray}{\pm0.004}}$ & \underline{0.980}$_{\textcolor{gray}{\pm0.002}}$ & 44 & 1.1$\times$ \\
    \bottomrule
\end{tabular}
  }
\end{table}

\subsection{Evaluation on Complex Dynamics}

Generating complex dynamics often exposes the fragility of the standard sampling trajectory. We evaluate the performance of \Ours against existing methods across the following three dimensions:
\noindent\textbf{Quantitative Comparison.} As summarized in Table\!~\ref{tab:main_results}, both variants of \Ours demonstrate consistent improvements in objective metrics across Dynamic-Bench and VideoJAM-Bench. On VideoJAM-Bench, \Ours-Adapter achieves the highest score in T2V Alignment (4.126), while \Ours-Zero achieves the highest Physical Consistency (3.448) and Visual Quality (3.939). On Dynamic-Bench, \Ours-Zero establishes the top T2V Alignment (3.651), while \Ours-Adapter achieves the highest Visual Quality (3.864). Compared to the baseline and other inference-time interventions (FlowMo and SRVS), our method effectively enhances Physical Consistency while maintaining highly competitive Motion Smoothness (\eg, reaching 0.980 on Dynamic-Bench). This indicates that amplitude rectification preserves the natural temporal progression of the video.

\noindent\textbf{Computational Efficiency.} \Ours is designed as an exceptionally lightweight, inference-time intervention. As shown in the Efficiency columns of Table\!~\ref{tab:main_results}, \Ours requires only 44 NFEs, introducing a minimal overhead (1.1$\times$ Time) over the standard 40-step baseline. This provides a highly favorable performance-compute trade-off compared to simply scaling up the ODE solver steps (NFE$\times$1.5), which demands 60 NFEs and a 1.5$\times$ time penalty. Furthermore, \Ours maintains a decisive efficiency advantage when benchmarked against other test-time techniques like SRVS (60 NFEs, 1.5$\times$ Time) and FlowMo (1.7$\times$ Time). See Appendix \S\ref{sec:supp_setup} for the memory overhead.

\begin{wrapfigure}[10]{r}{0.5\textwidth} 
  \centering
  \vspace{-15pt} 
  \includegraphics[width=\linewidth]{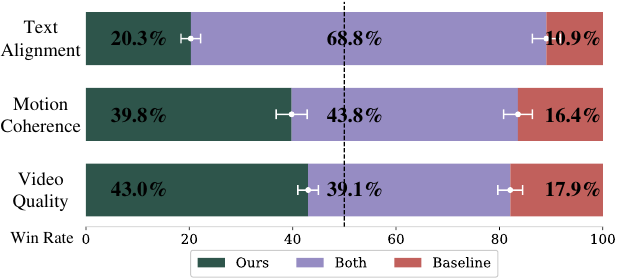}
  \caption{User study on VideoJAM-Bench.}
  \label{fig:user_study}
  \vspace{-10pt} 
\end{wrapfigure}

\noindent\textbf{User Study.} We conduct a blind human preference study on VideoJAM-Bench. Annotators evaluate randomized video pairs (the Wan2.2 baseline vs. \Ours) across three criteria: Text Alignment, Motion Coherence, and Video Quality. As illustrated in Fig.\!~\ref{fig:user_study}, \Ours consistently outperforms the baseline. Specifically, our method receives substantially higher preference votes in Video Quality (43.0\% vs. 17.9\%) and Motion Coherence (39.8\% vs. 16.4\%). Furthermore, while the robust baseline model already exhibits excellent capability in Text Alignment, resulting in a high tie rate of 68.8\%, \Ours still maintains a noticeable edge (20.3\% vs. 10.9\%). These results confirm that \Ours successfully translates into visually superior and motion-coherent dynamics, while fully preserving and even enhancing the base model's strong prompt adherence.

\begin{figure}[h!]
    \centering
    \includegraphics[width=\linewidth]{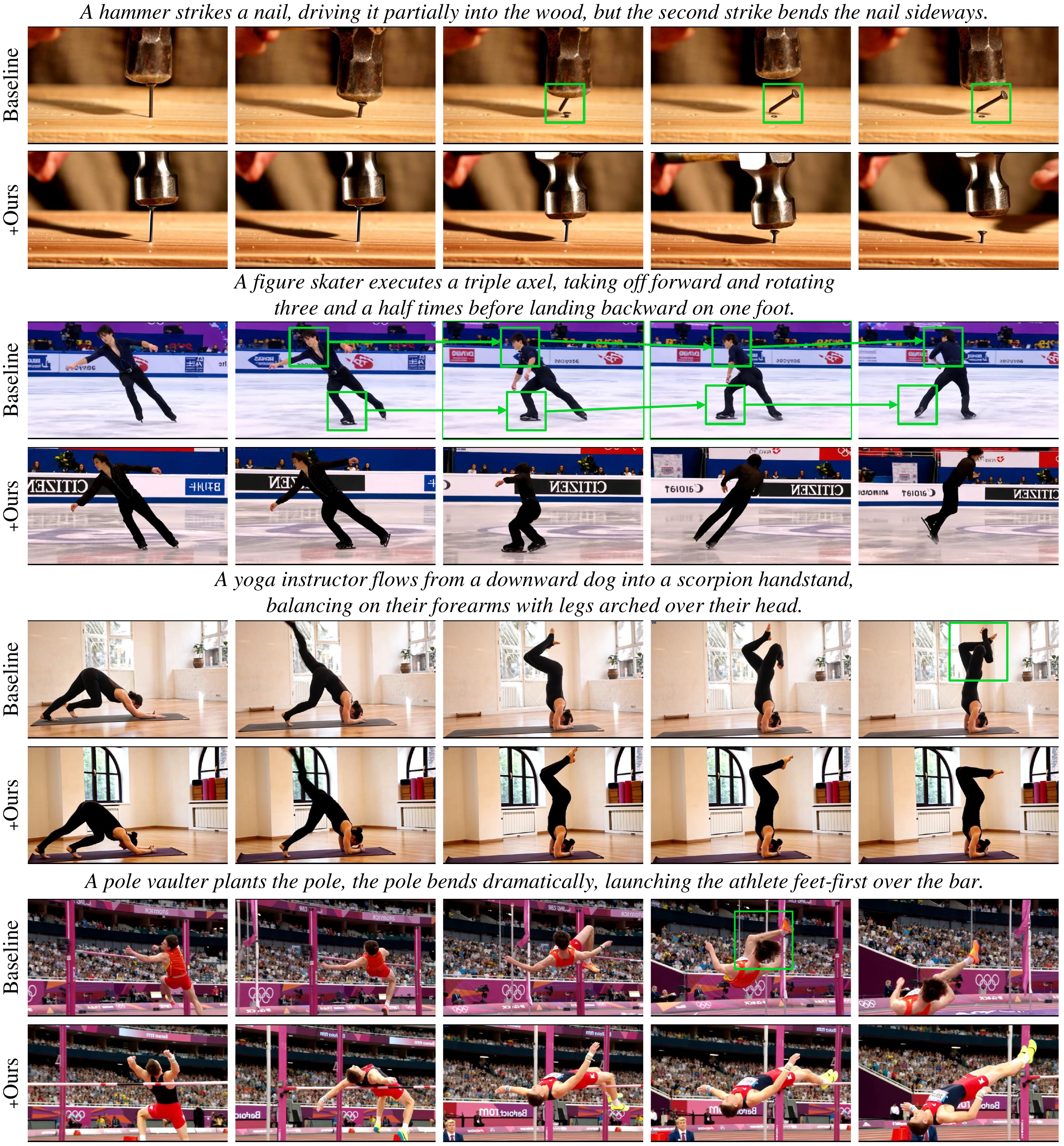}
    \caption{Qualitative comparison of spatially consistent video generation against Wan2.2-A14b T2V. Artifacts and structural errors are indicated by green bounding boxes.}
    \label{fig:qualitative}
    \vspace{-10pt}
\end{figure}

\noindent\textbf{Qualitative Results.} Visual inspections (as shown in Fig.\!~\ref{fig:qualitative}) corroborate our quantitative findings. While baseline trajectories and extended sampling approaches occasionally struggle with geometric distortions or semantic ambiguities under intense motion, \Ours yields clearer structural boundaries and fewer physically implausible artifacts. This explicitly aligns with the improved Physical Consistency metric, demonstrating that \Ours effectively mitigates trajectory drift.

\begin{figure}[t]
    \centering
    \includegraphics[width=\linewidth]{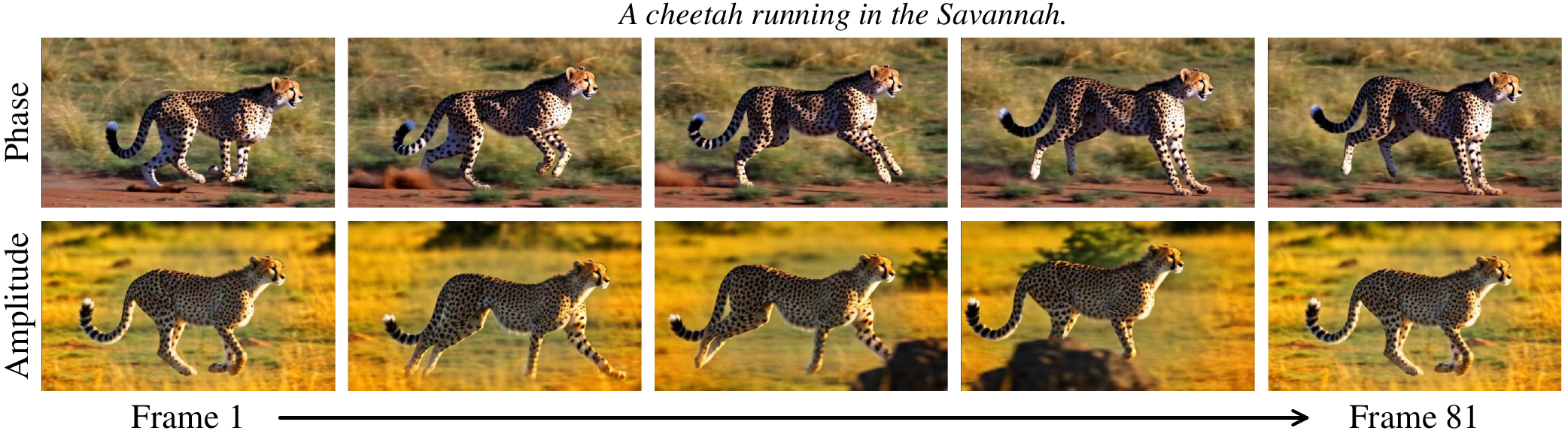}
    \caption{Diagnostic comparison of intervention targets (Phase \textit{vs.} Amplitude).}
    \label{fig:phase}
    \vspace{-10pt}
\end{figure}

\subsection{Diagnostic Experiment}
\label{subsec:diagnostic}
We validate our method via ablations on Wan2.2 using Dynamic-Bench. To isolate intervention effects, all diagnostics (except Table\!~\ref{tab:ablation_noise}) strictly employ Deterministic Continuation.

\noindent\textbf{Intervention Target: Amplitude vs. Phase.} We first investigate what happens if we intervene on the phase instead of the amplitude. We conducted experiments rectifying only the phase using global natural statistics. Rather than merely generating spatial artifacts, altering the instance-specific phase directly led to a fundamental breakdown of the underlying motion structure. As demonstrated in Fig.\!~\ref{fig:phase}, arbitrarily altering the phase disrupts delicate temporal alignment, causing severe dynamic anomalies (\eg, forcing fast-moving actions into disjointed, extremely slow-motion sequences). Because phase inherently encodes spatiotemporal displacement, its disruption primarily corrupts motion coherence without adding typical spatial noise. This strictly confirms that amplitude remains the only mathematically safe target for early-stage structural intervention. 

\noindent\textbf{The Role of Re-Noising Stochasticity.} Finally, we investigate the impact of the noise term $\epsilon$ during re-noising (Table\!~\ref{tab:ablation_noise}). Strict Deterministic Continuation ($\epsilon=\epsilon_{init}$) preserves the baseline's original kinetic energy, thereby retaining a higher Dynamic Degree while facilitating precisely controlled visual comparisons. Conversely, Stochastic Contraction ($\epsilon \sim \mathcal{N}(0, I)$) injects freshly sampled noise. Protected by the \Ours-rectified trajectory, this resets local discretization errors, yielding superior Consistency and Motion Smoothness. However, this reveals a fundamental consistency-dynamism trade-off, a phenomenon similarly observed in perturbation-based methods like SRVS$_{\!}$~\citep{jang2026self}. Stochastic noise acts as an implicit temporal regularizer that smooths the generative path; while this effectively suppresses structural artifacts, it inherently dampens high-frequency motion variance, deliberately trading peak dynamic intensity for global spatiotemporal stability.

\begin{table}[t!]
  \centering
    \begin{minipage}{0.49\linewidth}
    \centering
    \caption{Ablation on re-noising stochasticity.}
    \label{tab:ablation_noise}
    \resizebox{\linewidth}{!}{
    \begin{tabular}{l ccc}
      \toprule
      Setting & Consist. ($\uparrow$) & Smooth. ($\uparrow$) & Dyn. Deg. ($\uparrow$) \\
      \midrule
      Zero (Det.)   & 0.909 & 0.974 & \textbf{0.916} \\
      Zero (Stoch.) & \textbf{0.912} & \textbf{0.977} & 0.850 \\
      \midrule
      Adapt. (Det.) & 0.910 & 0.973 &\textbf{ 0.875} \\
      Adapt. (Stoch.)& \textbf{0.914} & \textbf{0.979} & 0.800 \\
      \bottomrule
    \end{tabular}
    }
    \end{minipage}\hfill
  \begin{minipage}{0.47\linewidth}
    \centering
    \caption{Ablation on modulation strength ($\lambda$).}
    \label{tab:ablation_lambda}
    \resizebox{\linewidth}{!}{
    \begin{tabular}{l ccc}
      \toprule
      Method & Consist. ($\uparrow$) & Smooth. ($\uparrow$) & Dyn. Deg. ($\uparrow$) \\
      \midrule
      $\lambda=1.0$ & 0.908 & 0.973 & 0.908 \\
      $\lambda=0.8$ & 0.909 & 0.973 & 0.883 \\
      $\lambda=0.5$ & \textbf{0.909} & \textbf{0.974} & \textbf{0.916} \\
      $\lambda=0.2$ & 0.910 & \textbf{0.974} & 0.900 \\
      \bottomrule
    \end{tabular}
    }
  \end{minipage}
  
\end{table}

\begin{figure}[ht!]
    \centering
    \includegraphics[width=\linewidth]{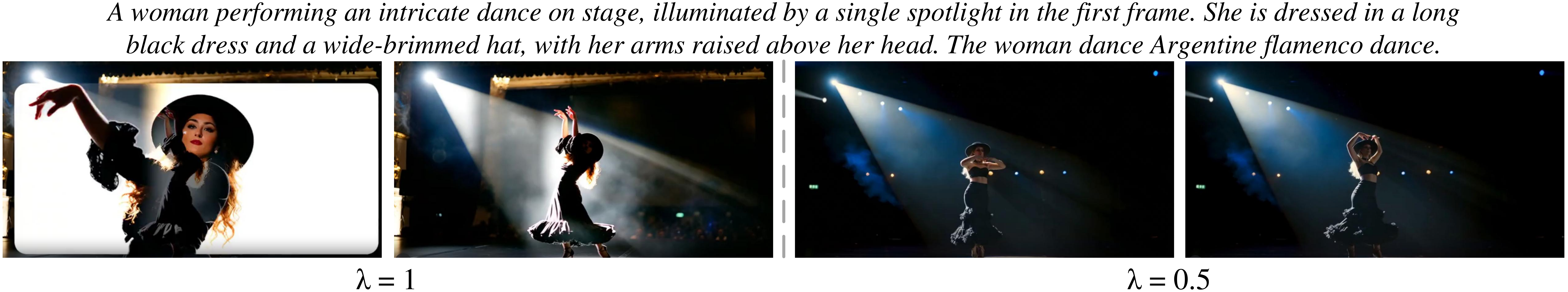}
    \caption{Impact of modulation strength ($\lambda$). An aggressive intervention with $\lambda=1.0$ (left) causes numerical instability, sometimes visually desynchronizing the initial frame from the sequence.}
    \label{fig:lambda}
\end{figure}

\noindent\textbf{Modulation Strength ($\lambda$) Dynamics.} As detailed in Table\!~\ref{tab:ablation_lambda}, we evaluated the intervention performance using $\lambda \in \{0.2, 0.5, 0.8, 1.0\}$. While overall quantitative differences are bounded, extreme values introduce specific qualitative failure modes. A weak intervention ($\lambda=0.2$) is insufficient to pull the trajectory out of the drift. Conversely, a highly aggressive replacement ($\lambda=1.0$) triggers numerical instability in subsequent ODE steps, causing the initial frame to become visually uncoordinated with the rest of the sequence (Fig.\!~\ref{fig:lambda}). Setting $\lambda=0.5$ emerges as the optimal sweet spot, providing a firm energy anchor while granting the ODE solver necessary mathematical flexibility.

\section{Discussion and Conclusion}

\noindent\textbf{Limitations.} While \Ours stabilizes structure by preserving the phase spectrum, this strict adherence can be a double-edged sword. If initial phase predictions are severely corrupted during the earliest ODE steps, locking the phase might trap extreme non-rigid transformations into unnatural states. Additionally, computing a global 3D spatiotemporal FFT introduces memory overhead that scales with spatial resolution and sequence length, and this formulation binds the pre-computed statistical prior to a fixed shape, necessitating a complete re-extraction if the resolution changes.

\noindent\textbf{Conclusion.} In this work, we address a fundamental bottleneck in Flow Matching-based video generation: the unchecked accumulation of trajectory drift during ODE integration, which irrecoverably corrupts spatiotemporal and structural coherence. To overcome this, we introduce Spectral Lookahead Rectification (\Ours), an efficient inference paradigm that shifts trajectory correction to the frequency domain. By diagnosing drift via lookahead projection, our method strategically anchors the macroscopic amplitude to match natural video priors during the earliest inference steps, while strictly protecting the geometry-encoding phase. This asymmetric treatment yields a critical insight: explicitly correcting amplitude organically drives the sensitive phase to converge toward structurally valid states. Consequently, this steers the generation trajectory back on track without relying on spatial perturbations. By resolving foundational trajectory drift with a minimal overhead of merely 4 additional NFEs, \Ours redefines the cost-performance trade-off in inference-time optimization. Ultimately, this mathematically sound, plug-and-play method boosts large-scale video foundation models, advancing reliable and motion-coherent spatiotemporal generation.

{\small
\bibliographystyle{unsrt}
\bibliography{main}

@String(CVPR  = {IEEE Conf. Comput. Vis. Pattern Recog.})

@String(ICCV  = {Int. Conf. Comput. Vis.})

@String(ECCV  = {Eur. Conf. Comput. Vis.})

@String(NeurIPS = {Adv. Neural Inform. Process. Syst.})

@String(ICML  = {Int. Conf. Mach. Learn.})

@String(ICLR  = {Int. Conf. Learn. Represent.})

@article{kong2024hunyuanvideo,
  title={{HunyuanVideo}: A systematic framework for large video generative models},
  author={Kong, Weijie and Tian, Qi and Zhang, Zijian and Min, Rox and Dai, Zuozhuo and Zhou, Jin and Xiong, Jiangfeng and Li, Xin and Wu, Bo and Zhang, Jianwei and others},
  journal={arXiv preprint arXiv:2412.03603},
  year={2024}
}

@article{wan2025wan,
  title={Wan: Open and advanced large-scale video generative models},
  author={Wan, Team and Wang, Ang and Ai, Baole and Wen, Bin and Mao, Chaojie and Xie, Chen-Wei and Chen, Di and Yu, Feiwu and Zhao, Haiming and Yang, Jianxiao and others},
  journal={arXiv preprint arXiv:2503.20314},
  year={2025}
}

@article{liu2024sora,
  title={Sora: A review on background, technology, limitations, and opportunities of large vision models},
  author={Liu, Yixin and Zhang, Kai and Li, Yuan and Yan, Zhiling and Gao, Chujie and Chen, Ruoxi and Yuan, Zhengqing and Huang, Yue and Sun, Hanchi and Gao, Jianfeng and others},
  journal={arXiv preprint arXiv:2402.17177},
  year={2024}
}

@article{team2025kling,
  title={{Kling-Omni} Technical Report},
  author={Team, Kling and Chen, Jialu and Ci, Yuanzheng and Du, Xiangyu and Feng, Zipeng and Gai, Kun and Guo, Sainan and Han, Feng and He, Jingbin and He, Kang and others},
  journal={arXiv preprint arXiv:2512.16776},
  year={2025}
}

@article{polyak2024movie,
  title={{Movie Gen}: A cast of media foundation models},
  author={Polyak, Adam and Zohar, Amit and Brown, Andrew and Tjandra, Andros and Sinha, Animesh and Lee, Ann and Vyas, Apoorv and Shi, Bowen and Ma, Chih-Yao and Chuang, Ching-Yao and others},
  journal={arXiv preprint arXiv:2410.13720},
  year={2024}
}

@inproceedings{villegasphenaki,
  title={Phenaki: Variable Length Video Generation from Open Domain Textual Descriptions},
  author={Villegas, Ruben and Babaeizadeh, Mohammad and Kindermans, Pieter-Jan and Moraldo, Hernan and Zhang, Han and Saffar, Mohammad Taghi and Castro, Santiago and Kunze, Julius and Erhan, Dumitru},
  booktitle=ICLR,
  year={2023}
}

@inproceedings{qiufreenoise,
  title={{FreeNoise}: Tuning-Free Longer Video Diffusion via Noise Rescheduling},
  author={Qiu, Haonan and Xia, Menghan and Zhang, Yong and He, Yingqing and Wang, Xintao and Shan, Ying and Liu, Ziwei},
  booktitle=ICLR,
  year={2024}
}

@inproceedings{lu2024freelong,
  title={{FreeLong}: Training-free long video generation with spectralblend temporal attention},
  author={Lu, Yu and Liang, Yuanzhi and Zhu, Linchao and Yang, Yi},
  booktitle=NeurIPS,
  pages={131434--131455},
  year={2024}
}

@inproceedings{li2025longdiff,
  title={{LongDiff}: Training-free long video generation in one go},
  author={Li, Zhuoling and Rahmani, Hossein and Ke, Qiuhong and Liu, Jun},
  booktitle=CVPR,
  pages={17789--17798},
  year={2025}
}

@inproceedings{lee2025videoguide,
  title={{VideoGuide}: Improving Video Diffusion Models without Training Through a Teacher's Guide},
  author={Lee, Dohun and Kim, Bryan Sangwoo and Park, Geon Yeong and Ye, Jong Chul},
  booktitle=CVPR,
  pages={2599--2608},
  year={2025}
}

@article{su2025theoretical,
  title={A theoretical analysis of discrete flow matching generative models},
  author={Su, Maojiang and Lu, Mingcheng and Hu, Jerry Yao-Chieh and Wu, Shang and Song, Zhao and Reneau, Alex and Liu, Han},
  journal={arXiv preprint arXiv:2509.22623},
  year={2025}
}

@article{gao2025seedance,
  title={Seedance 1.0: Exploring the boundaries of video generation models},
  author={Gao, Yu and Guo, Haoyuan and Hoang, Tuyen and Huang, Weilin and Jiang, Lu and Kong, Fangyuan and Li, Huixia and Li, Jiashi and Li, Liang and Li, Xiaojie and others},
  journal={arXiv preprint arXiv:2506.09113},
  year={2025}
}

@article{jang2026self,
  title={Self-Refining Video Sampling},
  author={Jang, Sangwon and Ki, Taekyung and Jo, Jaehyeong and Xie, Saining and Yoon, Jaehong and Hwang, Sung Ju},
  journal={arXiv preprint arXiv:2601.18577},
  year={2026}
}

@inproceedings{xu2023restart,
  title={Restart sampling for improving generative processes},
  author={Xu, Yilun and Deng, Mingyang and Cheng, Xiang and Tian, Yonglong and Liu, Ziming and Jaakkola, Tommi},
  booktitle=NeurIPS,
  pages={76806--76838},
  year={2023}
}

@inproceedings{na2024diffusion,
  title={Diffusion Rejection Sampling},
  author={Na, Byeonghu and Kim, Yeongmin and Park, Minsang and Shin, Donghyeok and Kang, Wanmo and Moon, Il Chul},
  booktitle=ICML,
  volume={235},
  pages={37097--37121},
  year={2024}
}

@inproceedings{rameshtest,
  title={Test-Time Scaling of Diffusion Models via Noise Trajectory Search},
  author={Ramesh, Vignav and Mardani, Morteza},
  booktitle=NeurIPS,
  year={2025}
}

@inproceedings{wu2024freeinit,
  title={{FreeInit}: Bridging initialization gap in video diffusion models},
  author={Wu, Tianxing and Si, Chenyang and Jiang, Yuming and Huang, Ziqi and Liu, Ziwei},
  booktitle=ECCV,
  pages={378--394},
  year={2024},
  organization={Springer}
}

@article{dong1995statistics,
  title={Statistics of natural time-varying images},
  author={Dong, Dawei W and Atick, Joseph J},
  journal={Network: computation in neural systems},
  volume={6},
  number={3},
  pages={345},
  year={1995},
}

@inproceedings{lipmanflow,
  title={Flow Matching for Generative Modeling},
  author={Lipman, Yaron and Chen, Ricky TQ and Ben-Hamu, Heli and Nickel, Maximilian and Le, Matthew},
  booktitle=ICLR,
  year={2023}
}

@inproceedings{peebles2023scalable,
  title={Scalable diffusion models with transformers},
  author={Peebles, William and Xie, Saining},
  booktitle=ICCV,
  pages={4195--4205},
  year={2023}
}

@inproceedings{ho2021classifier,
  title={Classifier-Free Diffusion Guidance},
  author={Ho, Jonathan and Salimans, Tim},
  booktitle={NeurIPS 2021 Workshop on Deep Generative Models and Downstream Applications},
  year=2021
}

@inproceedings{wu2025improved,
  title={Improved video vae for latent video diffusion model},
  author={Wu, Pingyu and Zhu, Kai and Liu, Yu and Zhao, Liming and Zhai, Wei and Cao, Yang and Zha, Zheng-Jun},
  booktitle=CVPR,
  pages={18124--18133},
  year={2025}
}

@inproceedings{albergo2023building,
  title={Building Normalizing Flows with Stochastic Interpolants},
  author={Albergo, Michael and Vanden-Eijnden, Eric},
  booktitle=ICLR,
  year={2023}
}

@inproceedings{liuflow,
  title={Flow Straight and Fast: Learning to Generate and Transfer Data with Rectified Flow},
  author={Liu, Xingchao and Gong, Chengyue and others},
  booktitle=ICLR,
  year={2023}
}

@inproceedings{singhalgeneral,
  title={A General Framework for Inference-time Scaling and Steering of Diffusion Models},
  author={Singhal, Raghav and Horvitz, Zachary and Teehan, Ryan and Ren, Mengye and Yu, Zhou and McKeown, Kathleen and Ranganath, Rajesh},
  booktitle=ICML,
  year={2024}
}

@inproceedings{oshimainference,
  title={Inference-Time Text-to-Video Alignment with Diffusion Latent Beam Search},
  author={Oshima, Yuta and Suzuki, Masahiro and Matsuo, Yutaka and Furuta, Hiroki},
  booktitle=NeurIPS,
  year={2025}
}

@article{he2025scaling,
  title={Scaling image and video generation via test-time evolutionary search},
  author={He, Haoran and Liang, Jiajun and Wang, Xintao and Wan, Pengfei and Zhang, Di and Gai, Kun and Pan, Ling},
  journal={arXiv preprint arXiv:2505.17618},
  year={2025}
}

@inproceedings{yuanfreqprior,
  title={{FreqPrior}: Improving Video Diffusion Models with Frequency Filtering Gaussian Noise},
  author={Yuan, Yunlong and Guo, Yuanfan and Wang, Chunwei and Zhang, Wei and Xu, Hang and Zhang, Li},
  booktitle=ICLR,
  year={2025}
}

@article{fan2025cfg,
  title={{CFG-Zero*}: Improved classifier-free guidance for flow matching models},
  author={Fan, Weichen and Zheng, Amber Yijia and Yeh, Raymond A and Liu, Ziwei},
  journal={arXiv preprint arXiv:2503.18886},
  year={2025}
}

@article{shaulov2025flowmo,
  title={{FlowMo}: Variance-based flow guidance for coherent motion in video generation},
  author={Shaulov, Ariel and Hazan, Itay and Wolf, Lior and Chefer, Hila},
  journal={arXiv preprint arXiv:2506.01144},
  year={2025}
}

@article{blattmann2023stable,
  title={Stable video diffusion: Scaling latent video diffusion models to large datasets},
  author={Blattmann, Andreas and Dockhorn, Tim and Kulal, Sumith and Mendelevitch, Daniel and Kilian, Maciej and Lorenz, Dominik and Levi, Yam and English, Zion and Voleti, Vikram and Letts, Adam and others},
  journal={arXiv preprint arXiv:2311.15127},
  year={2023}
}

@article{chefer2025videojam,
  title={{VideoJAM}: Joint appearance-motion representations for enhanced motion generation in video models},
  author={Chefer, Hila and Singer, Uriel and Zohar, Amit and Kirstain, Yuval and Polyak, Adam and Taigman, Yaniv and Wolf, Lior and Sheynin, Shelly},
  journal={arXiv preprint arXiv:2502.02492},
  year={2025}
}

@inproceedings{huang2024vbench,
  title={{VBench}: Comprehensive benchmark suite for video generative models},
  author={Huang, Ziqi and He, Yinan and Yu, Jiashuo and Zhang, Fan and Si, Chenyang and Jiang, Yuming and Zhang, Yuanhan and Wu, Tianxing and Jin, Qingyang and Chanpaisit, Nattapol and others},
  booktitle=CVPR,
  pages={21807--21818},
  year={2024}
}

@article{liu2025improving,
  title={Improving video generation with human feedback},
  author={Liu, Jie and Liu, Gongye and Liang, Jiajun and Yuan, Ziyang and Liu, Xiaokun and Zheng, Mingwu and Wu, Xiele and Wang, Qiulin and Xia, Menghan and Wang, Xintao and others},
  journal={arXiv preprint arXiv:2501.13918},
  year={2025}
}

@article{fang2025inflvg,
  title={{InfLVG}: Reinforce inference-time consistent long video generation with grpo},
  author={Fang, Xueji and Ma, Liyuan and Chen, Zhiyang and Zhou, Mingyuan and Qi, Guo-jun},
  journal={arXiv preprint arXiv:2505.17574},
  year={2025}
}

@inproceedings{nam2025optical,
  title={Optical-flow guided prompt optimization for coherent video generation},
  author={Nam, Hyelin and Kim, Jaemin and Lee, Dohun and Ye, Jong Chul},
  booktitle=CVPR,
  pages={7837--7846},
  year={2025}
}

@inproceedings{liu2025video,
  title={{Video-T1}: Test-time scaling for video generation},
  author={Liu, Fangfu and Wang, Hanyang and Cai, Yimo and Zhang, Kaiyan and Zhan, Xiaohang and Duan, Yueqi},
  booktitle=ICCV,
  pages={18671--18681},
  year={2025}
}

@article{luo2025enhance,
  title={{Enhance-A-Video}: Better generated video for free},
  author={Luo, Yang and Zhao, Xuanlei and Chen, Mengzhao and Zhang, Kaipeng and Shao, Wenqi and Wang, Kai and Wang, Zhangyang and You, Yang},
  journal={arXiv preprint arXiv:2502.07508},
  year={2025}
}

@article{chen2025temporal,
  title={Temporal regularization makes your video generator stronger},
  author={Chen, Harold Haodong and Huang, Haojian and Wu, Xianfeng and Liu, Yexin and Bai, Yajing and Shu, Wen-Jie and Yang, Harry and Lim, Ser-Nam},
  journal={arXiv preprint arXiv:2503.15417},
  year={2025}
}

@article{zhang2025waver,
  title={Waver: Wave your way to lifelike video generation},
  author={Zhang, Yifu and Yang, Hao and Zhang, Yuqi and Hu, Yifei and Zhu, Fengda and Lin, Chuang and Mei, Xiaofeng and Jiang, Yi and Peng, Bingyue and Yuan, Zehuan},
  journal={arXiv preprint arXiv:2508.15761},
  year={2025}
}

@inproceedings{
  chang2024how,
  title={How I Warped Your Noise: a Temporally-Correlated Noise Prior for Diffusion Models},
  author={Pascal Chang and Jingwei Tang and Markus Gross and Vinicius C. Azevedo},
  booktitle=ICLR,
  year={2024},
}

@article{hassan2025factorized,
  title={Factorized Video Generation: Decoupling Scene Construction and Temporal Synthesis in Text-to-Video Diffusion Models},
  author={Hassan, Mariam and Van Delft, Bastien and Li, Wuyang and Alahi, Alexandre},
  journal={arXiv preprint arXiv:2512.16371},
  year={2025}
}

@article{fei2025structure,
  title={Structure From Tracking: Distilling Structure-Preserving Motion for Video Generation},
  author={Fei, Yang and Stoica, George and Liu, Jingyuan and Chen, Qifeng and Krishna, Ranjay and Wang, Xiaojuan and Liu, Benlin},
  journal={arXiv preprint arXiv:2512.11792},
  year={2025}
}

@article{gokmen2025ropecraft,
  title={{RoPECraft}: Training-Free Motion Transfer with Trajectory-Guided RoPE Optimization on Diffusion Transformers},
  author={Gokmen, Ahmet Berke and Ekin, Yigit and Bilecen, Bahri Batuhan and Dundar, Aysegul},
  journal={arXiv preprint arXiv:2505.13344},
  year={2025}
}

@article{zhu2025motionrag,
  title={{MotionRAG}: Motion Retrieval-Augmented Image-to-Video Generation},
  author={Zhu, Chenhui and Wu, Yilu and Wang, Shuai and Wu, Gangshan and Wang, Limin},
  journal={arXiv preprint arXiv:2509.26391},
  year={2025}
}

@article{wang2023internvid,
  title={{InternVid}: A Large-scale Video-Text Dataset for Multimodal Understanding and Generation},
  author={Wang, Yi and He, Yinan and Li, Yizhuo and Li, Kunchang and Yu, Jiashuo and Ma, Xin and Chen, Xinyuan and Wang, Yaohui and Luo, Ping and Liu, Ziwei and Wang, Yali and Wang, Limin and Qiao, Yu},
  journal={arXiv preprint arXiv:2307.06942},
  year={2023}
}

@article{seed2026seedance2,
  title={Seedance 2.0: Advancing Video Generation for World Complexity},
  author={Team Seedance},
  journal={arXiv preprint arXiv:2604.14148},
  year={2026}
}

@inproceedings{ho2020denoising,
  title={Denoising diffusion probabilistic models},
  author={Ho, Jonathan and Jain, Ajay and Abbeel, Pieter},
  booktitle=NeurIPS,
  volume={33},
  pages={6840--6851},
  year={2020}
}

@inproceedings{rombach2022high,
  title={High-resolution image synthesis with latent diffusion models},
  author={Rombach, Robin and Blattmann, Andreas and Lorenz, Dominik and Esser, Patrick and Ommer, Bj{\"o}rn},
  booktitle=CVPR,
  pages={10684--10695},
  year={2022}
}

@inproceedings{zhao2023unipc,
  title={Unipc: A unified predictor-corrector framework for fast sampling of diffusion models},
  author={Zhao, Wenliang and Bai, Lujia and Rao, Yongming and Zhou, Jie and Lu, Jiwen},
  booktitle=NeurIPS,
  volume={36},
  pages={49842--49869},
  year={2023}
}

@article{oppenheim1981importance,
  title={The importance of phase in signals},
  author={Oppenheim, Alan V and Lim, Jae S},
  journal={Proceedings of the IEEE}, 
  year={1981},
  volume={69},
  number={5},
  pages={529-541}
}

@article{he2025videoscore2,
  title={{VideoScore2}: Think before you score in generative video evaluation},
  author={He, Xuan and Jiang, Dongfu and Nie, Ping and Liu, Minghao and Jiang, Zhengxuan and Su, Mingyi and Ma, Wentao and Lin, Junru and Ye, Chun and Lu, Yi and others},
  journal={arXiv preprint arXiv:2509.22799},
  year={2025}
}

@inproceedings{lv2024fourier,
  title={Fourier priors-guided diffusion for zero-shot joint low-light enhancement and deblurring},
  author={Lv, Xiaoqian and Zhang, Shengping and Wang, Chenyang and Zheng, Yichen and Zhong, Bineng and Li, Chongyi and Nie, Liqiang},
  booktitle=CVPR,
  pages={25378--25388},
  year={2024}
}

@inproceedings{si2024freeu,
  title={{FreeU}: Free lunch in diffusion u-net},
  author={Si, Chenyang and Huang, Ziqi and Jiang, Yuming and Liu, Ziwei},
  booktitle=CVPR,
  pages={4733--4743},
  year={2024}
}

@inproceedings{yang2025fam,
  title={{FAM Diffusion}: Frequency and attention modulation for high-resolution image generation with stable diffusion},
  author={Yang, Haosen and Bulat, Adrian and Hadji, Isma and Pham, Hai X and Zhu, Xiatian and Tzimiropoulos, Georgios and Martinez, Brais},
  booktitle=CVPR,
  pages={2459--2468},
  year={2025}
}

@article{wu2025hunyuanvideo,
  title={Hunyuanvideo 1.5 technical report},
  author={Wu, Bing and Zou, Chang and Li, Changlin and Huang, Duojun and Yang, Fang and Tan, Hao and Peng, Jack and Wu, Jianbing and Xiong, Jiangfeng and Jiang, Jie and others},
  journal={arXiv preprint arXiv:2511.18870},
  year={2025}
}
}

\newpage
\appendix
\renewcommand{\thetable}{S\arabic{table}}
\renewcommand{\thefigure}{S\arabic{figure}}
\setcounter{table}{0}
\setcounter{figure}{0}

\section*{A Summary of the Appendix}

We provide additional details and comprehensive analyses in this supplementary material, which are conceptually organized into three main parts:

\textbf{Part I: Methodological \& Theoretical Insights}
\begin{itemize}[leftmargin=*]
    \item \textbf{\S\ref{sec:supp_spectral}} provides an extended spectral analysis, empirically verifying that the spatiotemporal power-law prior is strictly preserved within the highly compressed 3D latent space.
    \item \textbf{\S\ref{sec:supp_discussion}} provides an extended discussion on related inference-time trajectory control methods, highlighting our conceptual and computational distinctions from stochastic spatial perturbation approaches.
\end{itemize}

\textbf{Part II: Reproducibility \& Implementation Details}
\begin{itemize}[leftmargin=*]
    \item \textbf{\S\ref{sec:supp_data}} elaborates on the data sourcing and code availability, providing a PyTorch-style pseudo-code to ensure immediate reproducibility of the core \Ours{} inference step.
    \item \textbf{\S\ref{sec:supp_licenses}} documents the licenses and terms of use for all pre-trained models, datasets, and evaluation benchmarks utilized in this research to ensure compliance.
    \item \textbf{\S\ref{sec:supp_setup}} introduces extended implementation details, including the global prior extraction process, the \Ours-Adapter architecture and training specifics, and an analysis of the memory overhead.
    \item \textbf{\S\ref{sec:user_study_details}} details the setup, double-blind protocol, and interface used for the human preference user study.
\end{itemize}

\textbf{Part III: Extended Experiments \& Applications}
\begin{itemize}[leftmargin=*]
    \item \textbf{\S\ref{sec:supp_more_exp}} presents extended experimental results and analyses. This section includes a sensitivity ablation on the Lookahead Window (\textbf{\S\ref{sec:lookaheadwindow}}), a discussion on method limitations via non-rigid/chaotic failure cases (\textbf{\S\ref{sec:supp_limitations}}), additional performance evaluations on high-fidelity cinematic scenarios (MovieGen-Bench$_{\!}$~\citep{polyak2024movie}) (\textbf{\S\ref{sec:moviegen}}), and a demonstration of our method's model-agnostic generalization to the \texttt{HunyuanVideo-1.5} architecture (\textbf{\S\ref{sec:hunyuan}}).
\end{itemize}

\begin{wrapfigure}[16]{r}{0.50\textwidth}
    \centering
    \vspace*{-10pt}
    \includegraphics[width=\linewidth]{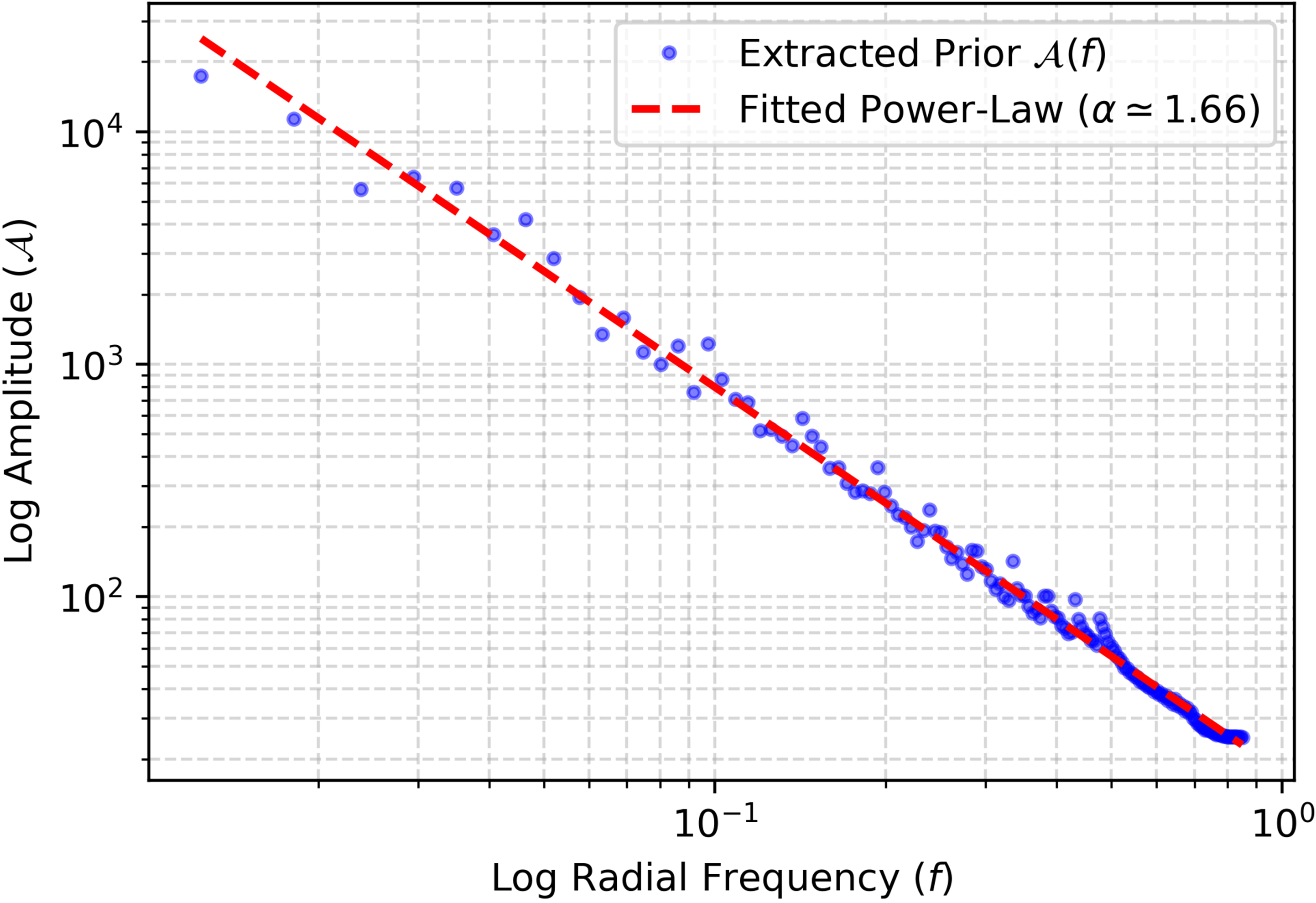}
    \caption{Empirical verification of the power-law prior in the 3D VAE latent space.}
    \label{fig:power_lawdecay}
\end{wrapfigure}

\section{Methodological \& Theoretical Insights}

\subsection{Empirical Verification of Spatiotemporal Power-Law} 
\label{sec:supp_spectral}
According to Natural Scene Statistics$_{\!}$~\citep{dong1995statistics}, real-world visual signals naturally exhibit a power-law decay in their amplitude spectrum. To empirically verify that this fundamental property is preserved within the highly compressed 3D latent space of modern video autoencoders, we aggregate the amplitude spectrum across 2,000 clean videos. Specifically, after computing the 3D FFT, we map the 3D Cartesian frequency coordinates $(u, v, w)$ to a 1D radial frequency $f = \sqrt{u^2 + v^2 + w^2}$. Here, $(u, v, w)$ explicitly represent the normalized digital frequencies obtained via FFT shifting, which are strictly bounded by the Nyquist limit $[-0.5, 0.5]$ across all dimensions. By averaging the amplitude values within discretized radial frequency bins, we collapse the 4D amplitude tensor $\mathcal{A} \in \mathbb{R}^{C \times T \times H \times W}$ into a simplified 1D radial amplitude profile $\mathcal{A}(f)$. As illustrated in Fig.\!~\ref{fig:power_lawdecay}, when plotted on a log-log scale, the extracted global prior strictly adheres to a linear decay, empirically confirming the theoretical relation $\mathcal{A}(f) \propto 1/f^\alpha$. Our linear regression yields an empirical spatiotemporal scaling exponent of $\alpha \simeq 1.66$.

\subsection{Extended Discussion on Trajectory Control}
\label{sec:supp_discussion}
Addressing trajectory drift in continuous-time generative models is a rapidly evolving area. Our proposed \Ours{} shares the foundational motivation of recent inference-time interventions, yet diverges significantly in its core mechanism.

\noindent\textbf{Stochastic Spatial Perturbation vs. Deterministic Spectral Rectification.}
Recent advancements, most notably Self-Refining Video Sampling (SRVS)$_{\!}$~\citep{jang2026self} and Restart Sampling$_{\!}$~\citep{xu2023restart}, tackle trajectory drift by injecting stochastic noise into intermediate spatial latents. These methods effectively "rewind" the trajectory to re-evaluate structural features and dilute accumulated numerical errors. While SRVS intelligently minimizes accumulated error to guide this perturbation, the stochastic nature of noise injection within the highly entangled spatial domain can still inadvertently disrupt fragile local geometry. In contrast, \Ours{} is strictly deterministic. By decoupling the signal via 3D FFT, we mathematically isolate the macro-drifting component (amplitude) from the structural blueprint (phase). This enables \Ours{} to explicitly correct the trajectory using a firm natural prior without introducing any unpredictable spatial variance.

\noindent\textbf{Spatial Optimization vs. Spectral Efficiency.} 
Other concurrent works, such as FlowMo$_{\!}$~\citep{shaulov2025flowmo}, enforce temporal coherence through variance-based guidance or gradient-based manipulation. While highly effective at maintaining motion consistency, spatial optimization often requires repeated forward passes or heavy gradient computations, substantially inflating the computational footprint (\eg, a $1.7\times$ increase in total inference time). \Ours{} fundamentally bypasses this bottleneck by leveraging the efficiency of spectral decoupling. Rectifying the amplitude spectrum via our lightweight \Ours-Zero introduces a mere 10\% overhead (4 additional NFEs), shifting the trajectory correction paradigm from computationally heavy spatial guidance to highly efficient spectral anchoring.

\section{Reproducibility \& Implementation Details}

\subsection{Data \& Code Availability}
\label{sec:supp_data}

\noindent\textbf{Data Sourcing.} Both the global prior extraction (\Ours-Zero) and the training of the context-aware adapter (\Ours-Adapter) utilize the exact same dataset subset, consisting of 2,000 randomly sampled high-quality video clips strictly sourced from the InternVid-10M-FLT dataset $_{\!}$~\citep{wang2023internvid}. 

\noindent\textbf{Code \& Reproducibility.} This supplementary material provides the source code for the core inference components of our proposed \Ours{} method. The complete codebase, encompassing the \Ours-Adapter training scripts, pre-calculated global prior statistics, and automated evaluation pipelines, will be open-sourced upon acceptance to facilitate future research in trajectory control for continuous-time generative models. To guarantee immediate reproducibility of our core contribution, we provide a PyTorch-style pseudo-code for the \Ours inference step in Algorithm \ref{alg:sslr_inference_pytorch}.

\subsection{Licenses of Assets Used}
\label{sec:supp_licenses}

To ensure full compliance with the NeurIPS guidelines regarding the use of existing assets, we document the licenses and terms of use for all pre-trained models, datasets, and evaluation benchmarks utilized in this research:

\begin{itemize}[leftmargin=*]
    \item \textbf{Wan2.2 Model Series:} The foundational text-to-video model, Wan2.2-A14B$_{\!}$~\citep{wan2025wan}, is distributed under the \textbf{Apache 2.0 License}, which permits free use, modification, and distribution for both research and commercial purposes.
    \item \textbf{InternVid-10M-FLT Dataset:} The 2,000 video clips used for extracting the global spectral prior and training the \Ours-Adapter were sampled from the InternVid dataset$_{\!}$~\citep{wang2023internvid}, which is available under the \textbf{CC BY-NC-SA 4.0
 License}.
    \item \textbf{Evaluation Benchmarks \& Metrics:} The standardized video evaluation frameworks and metric models employed in our study—including VBench$_{\!}$~\citep{huang2024vbench}, VideoScore2$_{\!}$~\citep{he2025videoscore2}, Dynamic-Bench$_{\!}$~\citep{jang2026self}, VideoJAM-Bench$_{\!}$~\citep{chefer2025videojam}, and MovieGen-Bench$_{\!}$~\citep{polyak2024movie}—are generally open-sourced under the \textbf{Apache 2.0 License}, \textbf{CC BY-NC 4.0
 License} or \textbf{MIT License} for research purposes.
\end{itemize}

We confirm that our usage of these assets is strictly limited to research and evaluation, fully respecting the original creators' terms of service and copyright notices.

\begin{algorithm}[h!]
\renewcommand\thealgorithm{S1}
\caption{Pseudo-code for the \Ours Inference Step in a PyTorch-like style.}
\label{alg:sslr_inference_pytorch}
\lstset{
  backgroundcolor=\color{white},
  basicstyle=\fontsize{8pt}{9pt}\ttfamily\selectfont,
  columns=fullflexible,
  breaklines=true,
  captionpos=b,
  escapeinside={(:}{:)},
  commentstyle=\color{codecomment},
  keywordstyle=\color{codedefine}\textbf,
}
\begin{lstlisting}[language=python]
# Models & Variables:
# v_theta: Flow Matching velocity network (\eg, Wan2.2 DiT).
# freq_adapter: Optional trained DiT for context-aware residual amplitude.
# initial_noise: The exact initial Gaussian noise map sampled at t=1.
(:\color{codedefine}{\textbf{def}}:) (:\color{codefunc}{\textbf{speclor\_step}}:)(z_t, t, text_emb, v_theta, freq_adapter, global_prior, initial_noise, lambda_weight=0.5):
    # 1. Lookahead Projection (Eq. 1)
    velocity = (:\color{codefunc}{\textbf{v\_theta}}:)(z_t, t)
    z_t0_pred = z_t - t * velocity 
    
    # 2. Spatiotemporal Frequency Decoupling via 3D FFT
    # z_t0_pred shape: [B, C, T, H, W]
    fft_z = (:\color{codefunc}{\textbf{torch.fft.fftn}}:)(z_t0_pred.float(), dim=(-3, -2, -1))
    amplitude = (:\color{codefunc}{\textbf{torch.abs}}:)(fft_z)
    phase = (:\color{codefunc}{\textbf{torch.angle}}:)(fft_z)
    
    # 3. Amplitude Rectification
    (:\color{codedefine}{\textbf{if}}:) freq_adapter (:\color{codedefine}{\textbf{is not None}}:): # SpecLoR-Adapter (Option B)
        delta_A = (:\color{codefunc}{\textbf{freq\_adapter}}:)(z_t0_pred, t, text_emb)
        A_rectified = amplitude + delta_A
    (:\color{codedefine}{\textbf{else}}:): # SpecLoR-Zero Global Prior (Option A)
        # Note: mu_ref and sigma_ref are full 4D Tensors [C, T, H, W]
        mu_ref, sigma_ref = global_prior 
        # Apply asymmetric trust region clipping (Eq. 2)
        gamma, k = 0.6, 3.0
        A_target = (:\color{codefunc}{\textbf{torch.clamp}}:)(amplitude, min=gamma * mu_ref, max=mu_ref + k * sigma_ref)
        # Apply bounded residual update (Eq. 3)
        A_rectified = amplitude + lambda_weight * (A_target - amplitude)

    # 4. Re-Noising & Inference
    # Reconstruct the spectrum keeping Phase strictly locked
    fft_rectified = A_rectified * (:\color{codefunc}{\textbf{torch.exp}}:)(1j * phase)
    z_t0_rectified = (:\color{codefunc}{\textbf{torch.fft.ifftn}}:)(fft_rectified, dim=(-3, -2, -1)).real
    
    # Anchor the trajectory back to the current timestep (Eq. 5)
    z_t_rectified = (1 - t) * z_t0_rectified + t * initial_noise
    
    (:\color{codedefine}{\textbf{return}}:) z_t_rectified
\end{lstlisting}
\end{algorithm}

\subsection{Extended Implementation Details}
\label{sec:supp_setup}

\noindent\textbf{Global Prior Statistics (\Ours-Zero).} The 2,000 sampled videos from InternVid-10M-FLT are preprocessed to a spatial resolution of $480 \times 832$ with a sequence length of $81$ frames. We encode these clean videos using the frozen Variational Autoencoder (VAE) of Wan2.2-A14B $_{\!}$~\citep{wan2025wan}. To ensure numerical precision during frequency decoupling, latent representations are explicitly cast to \texttt{float32} before the 3D FFT. Crucially, rather than averaging the spectrum over spatiotemporal dimensions to obtain a single scalar per channel, we preserve the entire structural blueprint. Thus, the extracted global prior mean $\mu_{ref}$ and standard deviation $\sigma_{ref}$ are highly detailed 4D tensors of shape $C \times T \times H \times W$ (specifically, $16 \times 21 \times 60 \times 104$).

\noindent\textbf{\Ours-Adapter Architecture.} To capture context-aware dynamic variance, \Ours-Adapter is designed as a highly lightweight architecture constructed from merely 2 Diffusion Transformer (DiT) $_{\!}$~\citep{peebles2023scalable} blocks. Instead of processing spatial latents directly, the adapter performs a 3D FFT on the lookahead latent $z_{t,0}$. The resulting magnitude spectrum undergoes a logarithmic mapping, $\log(1 + \mathcal{A})$, before projection into a compact hidden dimension of 128. Rotary Positional Embeddings (RoPE) are explicitly applied along the frequency dimension. Cross-attention layers within each 4-head DiT block condition the residual amplitude prediction on text embeddings (context dimension of 4096). We employ the AdamW optimizer (learning rate $5\!\times\!10^{-5}$, cosine scheduler, 300 warmup steps) to minimize the MSE between predicted and ground-truth logarithmic magnitudes. The model is trained for 30 epochs (batch size 1, 4 gradient accumulation steps) in \texttt{bfloat16} precision, while the foundational Wan2.2 parameters remain strictly frozen.

\noindent\textbf{Memory Overhead of Spatiotemporal FFT.} While \Ours achieves exceptional computational efficiency regarding NFEs, computing the full 3D FFT via \texttt{torch.fft.fftn} introduces a memory footprint during inference. For a 14B model like Wan2.2, standard latents ($z_{t,0}$) are in 16-bit float (2 bytes/element). However, FFT mathematical operations require PyTorch to upcast these real tensors to \texttt{complex64} (32-bit real + 32-bit imaginary, totaling 8 bytes/element). Given a batch size $B$, channels $C$, and spatiotemporal dimensions $T \times H \times W$, the peak memory footprint for the transformed tensor $\mathcal{A}_{curr}$ becomes:
\begin{equation}
M_{FFT} = B \times C \times T \times H \times W \times 8 \text{ bytes}.
\end{equation}
While perfectly manageable for standard resolutions (\eg, $480 \times 832$, 81 frames), this quadratic scaling highlights a potential bottleneck for future ultra-high-resolution or minutes-long video generation, necessitating future explorations into windowed or patch-based spectral approximations.

\subsection{Details of the User Study}
\label{sec:user_study_details}

\begin{figure}[t!]
  \centering
  \includegraphics[width=\linewidth]{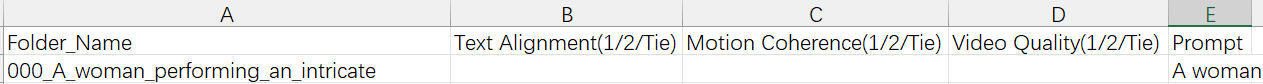}
  \caption{An example of the local scoring spreadsheet used by evaluators to record their preferences across three criteria for each anonymized video pair.}
  \label{fig:script}
\end{figure}

Five graduate student volunteers with computer vision backgrounds conducted the evaluation. To ensure a strict double-blind setup, we employed a local evaluation protocol. A custom Python script randomly shuffled and renamed the video pairs (Baseline vs. \Ours) to eliminate positional bias. Evaluators viewed these anonymized pairs locally and recorded their preferences (Video 1, Video 2, or Tie) across Text Alignment, Motion Coherence, and Video Quality, as illustrated in Fig.\!~\ref{fig:script}. The randomization script is provided in the supplementary materials.

\section{Extended Experiments \& Applications}
\label{sec:supp_more_exp}

\begin{table}[htbp]
\centering
\caption{Quantitative evaluation of intervention window timing on VideoJAM-Bench. Early intervention (Steps 2--5) is crucial; late interventions fail to rectify already-corrupted phase geometry.}
\label{tab:window_ablation}
\resizebox{\linewidth}{!}{
\begin{tabular}{l ccccc}
\toprule
Method & Consistency ($\uparrow$) & Motion Smooth. ($\uparrow$) & Dynamic Degree ($\uparrow$) & NFE & Time \\
\midrule
Wan2.2 T2V (UniPC)           & 0.935 & 0.972 & \textbf{0.875}  & 40 & \\
\midrule
+ \Ours-Zero (step 6-9)             & 0.935 & 0.971 & 0.844  & 44 & 1.1$\times$ \\
+ \Ours-Zero (step 10-13)           & 0.933 & 0.971 & 0.844  & 44 & 1.1$\times$ \\
+ \Ours-Zero (step 14-17)           & 0.932 & 0.970 & 0.856  & 44 & 1.1$\times$ \\
\midrule
+ \Ours-Zero (step 2-5)             & 0.939 & 0.973 & 0.859  & 44 & 1.1$\times$ \\
+ \Ours-Adapter (step 2-5)          & \textbf{0.942} & \textbf{0.978} & 0.783 & 44 & 1.1$\times$ \\
\bottomrule
\end{tabular}
}
\end{table}

\subsection{Sensitivity to the Lookahead Window} 
\label{sec:lookaheadwindow}
In the main text, we rigidly restrict the \Ours~intervention to steps 2 to 5 of a 40-step UniPC schedule. Table\!~\ref{tab:window_ablation} empirically demonstrates why isolating this specific early, high-noise regime is mathematically critical. 

\noindent\textbf{The Cost of Late Intervention.} Delaying the rectification window to later stages (\eg, steps 10-13 or 14-17) completely fails to rescue the trajectory. As shown, Subject Consistency progressively deteriorates to 0.933 and 0.932, while Motion Smoothness falls to 0.971 and 0.970, performing demonstrably worse than the unguided baseline. By step 6 and beyond, continuous trajectory drift has already permanently corrupted the geometric phase. Attempting to repair the amplitude at these later stages cannot undo established structural tearing (\eg, duplicated limbs). 

\noindent\textbf{The Early Advantage.} Conversely, anchoring the amplitude specifically during steps 2-5 successfully guides the trajectory before severe structural drift solidifies, achieving highly competitive Consistency (0.939) and Motion Smoothness (0.973) with the zero-cost prior, and peaking at 0.942 and 0.978 using our \Ours-Adapter.

\subsection{Limitations: Chaotic and Non-Rigid Transformations}
\label{sec:supp_limitations}

\begin{figure}[t!]
  \centering
   \includegraphics[width=\linewidth]{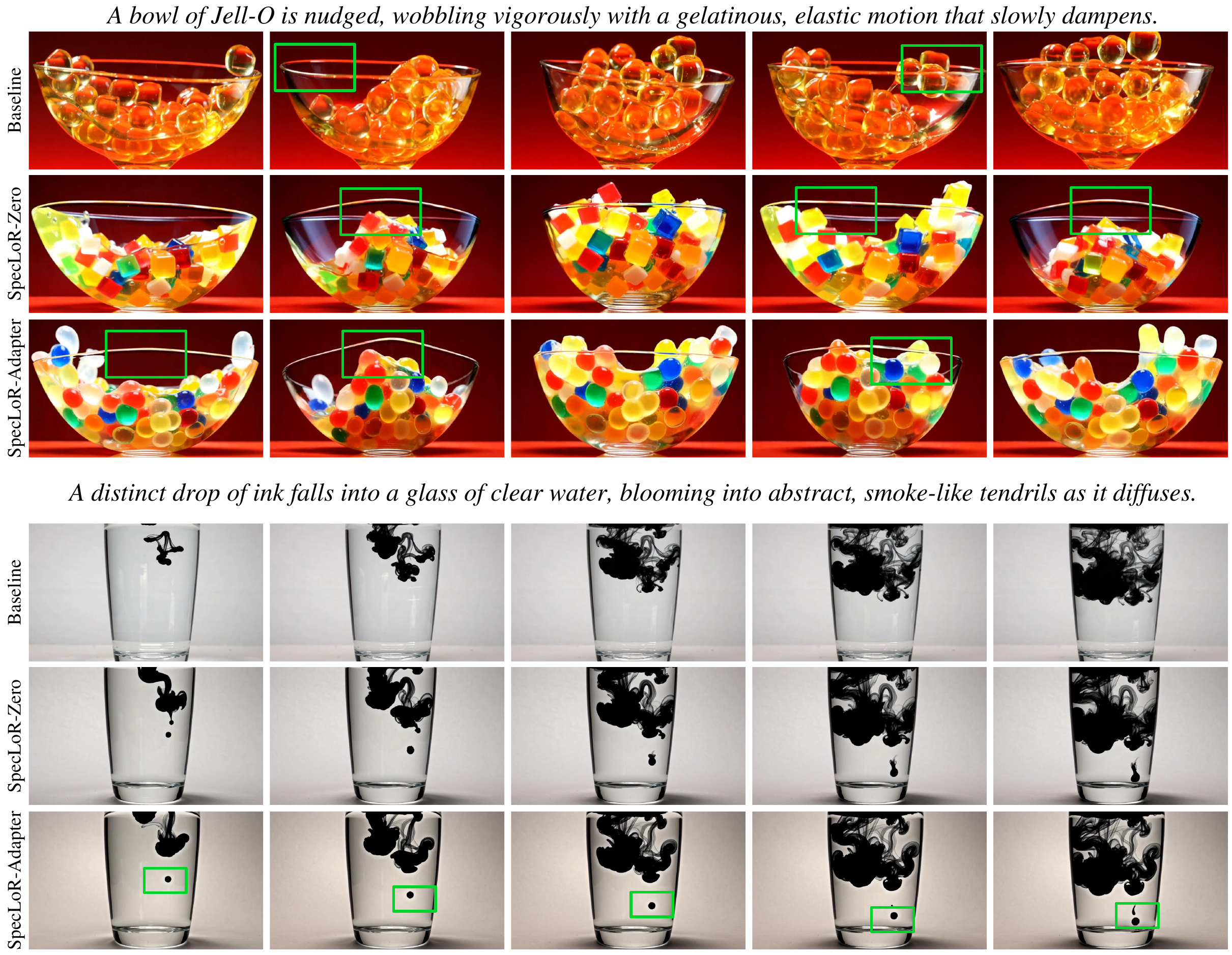}
  \caption{Visual demonstration of failure cases under chaotic, non-rigid transformations. By strictly preserving the early-stage phase, \Ours~can inadvertently lock the generation into an unnatural geometric state, such as transferring gelatinous properties to a rigid glass bowl (top) or preventing the natural topological diffusion of an ink drop (bottom).}
  \label{fig:failure_case}
\end{figure}

While \Ours excels at preserving kinematic structure for rigid bodies and human dynamics by locking the phase spectrum, this strict adherence introduces failure modes in highly chaotic systems. Fig.\!~\ref{fig:failure_case} visually dissects these limitations under extreme non-rigid transformations.

\noindent\textbf{Elastic Dynamics (Jell-O).} When evaluating gelatinous, wobbly motion, leaving the chaotic phase dynamics unmodified causes physical properties to erroneously bleed into the environment. As shown in the top row of Fig.\!~\ref{fig:failure_case}, the rigid glass bowl itself begins to drastically warp and wobble in the \Ours output, adopting the elastic properties of the Jell-O.

\noindent\textbf{Fluid Diffusion (Ink).} In scenarios involving rapid topological changes, such as an ink drop diffusing in water, strictly preserving the early-stage phase inadvertently locks the generation into an unnatural geometric state. Rather than blooming into smoke-like tendrils, the ink droplet persists as an unnatural, rigid spherical structure falling through the water (bottom row, green boxes), fundamentally failing to capture the continuous destruction of diffusion. This suggests that future trajectory control mechanisms must distinguish between rigid structural preservation and non-rigid chaotic evolution.

We empirically observe that employing the stochastic variant of our method partially mitigates this limitation, as the injected noise provides the necessary geometric flexibility to break these unnatural rigid locks. This suggests that future trajectory control mechanisms must distinguish between rigid structural preservation and non-rigid chaotic evolution.

\subsection{Motion Coherence under High-Fidelity Cinematic Scenarios}
\label{sec:moviegen}

\begin{table}[htbp]
\centering
\caption{Quantitative results on MovieGen-Bench. Both variants of \Ours substantially lift the Dynamic Degree and Motion Smoothness with minimal computational overhead, proving strong generalization to high-fidelity cinematic generations.}
\label{tab:moviegen_bench}
\resizebox{\linewidth}{!}{
\begin{tabular}{l ccccc}
\toprule
Method & Consistency ($\uparrow$) & Motion Smooth. ($\uparrow$) & Dynamic Degree ($\uparrow$) & NFE & Time \\
\midrule
Wan2.2 T2V (UniPC)           & 0.932 & 0.977 & 0.754 & 40 &  \\
+ NFE$\times$1.5             & 0.932 & 0.977 & 0.757 & 60 & 1.5$\times$ \\
\midrule
+ \textbf{\Ours-Zero (Ours)}          & \textbf{0.933} & \textbf{0.978} & 0.782 & 44 & 1.1$\times$ \\
+ \textbf{\Ours-Adapter (Ours)}       & 0.932 & \textbf{0.978} & \textbf{0.783} & 44 & 1.1$\times$ \\
\bottomrule
\end{tabular}
}
\end{table}

Beyond structural physics in local interactions, we further investigated whether amplitude rectification improves overall motion stability in high-fidelity, high-resolution cinematic generations. As presented in Table\!~\ref{tab:moviegen_bench}, both \Ours-Zero and \Ours-Adapter exhibit substantial gains on the rigorous MovieGen-Bench$_{\!}$~\citep{polyak2024movie}, comprehensively lifting the Dynamic Degree and Motion Smoothness without sacrificing base Consistency. Because \Ours operates strictly in the frequency domain and safely rectifies only the macroscopic energy, it intrinsically avoids injecting conflicting spatial patterns that could destabilize camera panning or zooming, a common vulnerability in stochastic perturbation methods. This mechanism ensures fluid subject motion and highly stable background consistency. The context-aware \Ours-Adapter further pushes this ceiling, optimizing complex motion trajectories to achieve peak dynamic performance while maintaining strong prompt adherence, ensuring high-fidelity scenes remain both dynamic and coherent.

\subsection{Generalization to HunyuanVideo-1.5}
\label{sec:hunyuan}

To verify the model-agnostic nature of \Ours, we extend the framework to the \texttt{HunyuanVideo-1.5} (480p) architecture$_{\!}$~\citep{wu2025hunyuanvideo}. Due to time and computational constraints, we focus our evaluation in this section exclusively on the zero-cost \texttt{SpecLoR-Zero} variant.

Furthermore, during inference, we empirically adjusted the default Classifier-Free Guidance (CFG) scale down to 3.0. While the baseline \texttt{HunyuanVideo} model often operates at higher guidance scales (e.g., 6.0), applying our strict spectral amplitude rectification under such aggressive CFG introduces severe visual artifacts. This occurs because an overly high CFG actively pushes latents into saturated, high-variance regions, creating a mathematical conflict with \texttt{SpecLoR-Zero}'s objective of anchoring macroscopic energy back to a natural prior. By lowering the CFG to 3.0, we effectively harmonize the text guidance with our spectral regularization, yielding coherent and high-fidelity dynamics.

For the 50-step sampling schedule used in HunyuanVideo-1.5, we set the Lookahead Window to steps 4--10. This configuration is chosen for two primary reasons:
\begin{itemize}[nosep,leftmargin=*]
    \item \textbf{Efficiency Consistency:} Applying 7 additional steps in a 50-step schedule incurs a modest $14\%$ computational overhead, keeping the relative cost comparable to our 40-step Wan2.2 experiments.
    \item \textbf{Timestep Alignment:} In the discrete flow-matching process, steps 4--10 in a 50-step schedule correspond to a noise-level range (timesteps) that is more equivalent to steps 2--5 in a 40-step schedule. This ensures that the spectral rectification targets the same critical early stage where global motion structures begin to coalesce.
\end{itemize}

Preliminary results (See Table\!~\ref{tab:hunyuan_results}) confirm that \Ours successfully suppresses structural artifacts and improves motion stability in HunyuanVideo-1.5 without architecture-specific tuning.

\begin{table}[t]
  \centering
  \caption{Quantitative comparison on \texttt{HunyuanVideo-1.5}. Best scores are in \textbf{bold}.}
  \label{tab:hunyuan_results}
  \resizebox{\textwidth}{!}{
  \begin{tabular}{lccc cc cc}
    \toprule
    & \multicolumn{3}{c}{VideoScore2} & \multicolumn{2}{c}{VBench} & \multicolumn{2}{c}{Efficiency} \\
    \cmidrule(r){2-4} \cmidrule(lr){5-6} \cmidrule(l){7-8}
    Method & T2V Align.$\uparrow$ & Phys. Cons.$\uparrow$ & Visual Qual.$\uparrow$ & Consistency$\uparrow$ & Motion Smooth.$\uparrow$ & NFE$\downarrow$ & Time$\downarrow$ \\
    \midrule
    \multicolumn{8}{c}{\textit{Results on VideoJAM-Bench$_{\!}$~\citep{chefer2025videojam}}} \\
    \midrule
    HunyuanVideo-1.5 & 3.833 & 2.924 & 3.486 & 0.938 & 0.985 & 50  &  \\
    + \Ours-Zero (Ours) & \textbf{3.890}$_{\textcolor{gray}{\pm0.015}}$ & \textbf{3.260}$_{\textcolor{gray}{\pm0.017}}$ & \textbf{3.736}$_{\textcolor{gray}{\pm0.012}}$ & \textbf{0.943}$_{\textcolor{gray}{\pm0.003}}$ & \textbf{0.992}$_{\textcolor{gray}{\pm0.002}}$ & 57 & 1.14$\times$ \\
    \midrule
    \multicolumn{8}{c}{\textit{Results on Dynamic-Bench$_{\!}$~\citep{jang2026self}}} \\
    \midrule
    HunyuanVideo-1.5 & 3.213 & 2.634 & 3.532 & 0.911 & 0.987 & 50  &  \\
    + \Ours-Zero (Ours) & \textbf{3.283}$_{\textcolor{gray}{\pm0.011}}$ & \textbf{2.785}$_{\textcolor{gray}{\pm0.018}}$ & \textbf{3.674}$_{\textcolor{gray}{\pm0.014}}$ & \textbf{0.915}$_{\textcolor{gray}{\pm0.004}}$ & \textbf{0.993}$_{\textcolor{gray}{\pm0.001}}$ & 57 & 1.14$\times$ \\
    \bottomrule
  \end{tabular}
  }
\end{table}

\end{document}